\documentclass[journal]{IEEEtran}
\usepackage{amsfonts}
\usepackage{upgreek}
\usepackage[ruled,vlined]{algorithm2e}
\usepackage{makecell,multirow,diagbox}
\usepackage{graphicx}
\usepackage{subfigure}
\usepackage{cite}
\usepackage{amsmath,bm}
\usepackage{accents}
\usepackage{amssymb}
\usepackage{amsmath}
\usepackage{graphicx}
\usepackage{epstopdf}
\usepackage{subfigure}
\usepackage{multirow}
\usepackage{makecell}
\usepackage{diagbox}
\usepackage{color}
\usepackage{tablor}
\usepackage{array}
\usepackage{url}
\usepackage[ruled,vlined]{algorithm2e}
\ifCLASSINFOpdf
\else
\fi

\begin{document}
%%%%%%%%%%%%%%%%%%%%%%%%%%%%%%%%%%%%%%%%%%%%%%%%%%%%%%%%%%%
\title{Hyperspectral Image Classification with Spatial Consistence Using Fully Convolutional Spatial Propagation Network\thanks{Y. Jiang and S. Zou are with the School of Computer Science, National Engineering Laboratory for Integrated Aero-Space-Ground-Ocean Big Data Application Technology, Shaanxi Provincial Key Laboratory of Speech and Image Information Processing, Northwestern Polytechnical University, Xi'an 710129, China (e-mail: ynjiang@mail.nwpu.edu.cn; shanrong.zou@outlook.com).} \thanks{H. Zhang is with the School of Computer Science, National Engineering Laboratory for Integrated Aero-Space-Ground-Ocean Big Data Application Technology, Shaanxi Provincial Key Laboratory of Speech and Image Information Processing, Northwestern Polytechnical University, Xi'an, China 710072, China, and also with the School of Computer Science, The University of Adelaide, Adelaide, SA 5005, Australia (e-mail: hkzhang1991@mail.nwpu.edu.cn).} \thanks{Y. Ling is with the School of Computer Science, National Engineering Laboratory for Integrated Aero-Space-Ground-Ocean Big Data Application Technology, Shaanxi Provincial Key Laboratory of Speech and Image Information Processing, Northwestern Polytechnical University, Xi'an, China 710072, China, and also with the National Key Laboratory of Science and Technology on Space Microwave, Xi'an 710000, China (e-mail: lybyp@nwpu.edu.cn).}\thanks{Y. Bai is with the School of Computing and Information Systems, The University of Melbourne, Melbourne, VIC 3010, Australia (e-mail: yunpengb@student.unimelb.edu.au).}} 

\author{Yenan Jiang, Ying Li, Shanrong Zou, Haokui Zhang, Yunpeng Bai}% <-this % stops a space

\markboth{SUBMITTED TO IEEE TRANSACTIONS ON GEOSCIENCE AND REMOTE SENSING}%
{Shell \MakeLowercase{\textit{et al.}}: Bare Demo of IEEEtran.cls for IEEE Journals}

\maketitle

\begin{abstract}
In recent years, deep convolutional neural networks (CNNs) have demonstrated impressive ability to represent hyperspectral images (HSIs) and achieved encouraging results in HSI classification.
%However, the traditional CNN models can only operate convolution on regular square image regions with local patches and fixed size, which causes failures in modeling contextual spatial information.
%However, the existing CNN-based models operate at the patch-level rather than the pixel-level.
%Especially in the case of overlapping image patches, since they are considered as independent samples, which will result in significant redundancy in computation and storage. 
However, the existing CNN-based models operate at the patch-level, in which pixel is separately classified into classes using a patch of images around it. 
This patch-level classification will lead to a large number of repeated calculations, and it is hard to identify the appropriate patch size that is beneficial to classification accuracy.    
%In addition, a local patch is utilized for each pixel to execute the training and prediction, which will lead to a large number of repeated calculations, and it is difficult to determine the appropriate patch size that is beneficial to classification accuracy.  
%In addition, the conventional CNN models operate convolutions with local receptive fields, which cause failures in modeling contextual spatial information.
In addition, the conventional CNN models operate convolutions with local receptive fields, which cause the failure of contextual spatial information modeling.
To overcome these aforementioned limitations, we propose a novel end-to-end, pixel-to-pixel fully convolutional spatial propagation network (FCSPN) for HSI classification. 
Our FCSPN consists of a 3D fully convolution network (3D-FCN) and a convolutional spatial propagation network (CSPN). 
%has the ability to deal with the arbitrary size of the input HSI adaptively, 
Specifically, the 3D-FCN is firstly introduced for reliable preliminary classification, in which a novel dual separable residual (DSR) unit is proposed to effectively capture spectral and spatial information simultaneously with fewer parameters.
Moreover, the channel-wise attention mechanism is adapted in the 3D-FCN to grasp the most informative channels from redundant channel information.
%Specifically, with the 3D-FCN a reliable pre-classification result can be gained, in which a novel dual separable residual (DSR) unit is presented so that spectral and spatial information can be efficiently captured simultaneously. Meanwhile, the attention mechanism is adapted to grasp the most robust spectral features from redundant spectral information.
Finally, the CSPN is introduced to capture the spatial correlations of HSI via learning a local linear spatial propagation, which allows maintaining the HSI spatial consistency and further refining the classification results. 
Experimental results on three HSI benchmark datasets demonstrate that the proposed FCSPN achieves state-of-the-art performance on HSI classification.

\end{abstract}

\begin{IEEEkeywords}
deep learning (DL), hyperspectral image (HSI) classification, 3D fully convolution network (3D-FCN), convolutional spatial propagation network (CSPN), attention.
\end{IEEEkeywords}

\IEEEpeerreviewmaketitle

\section{Introduction}
\IEEEPARstart{H}{yperspectral} image (HSI) has become a brisk research field due to its advanced abilities in the analysis of the Earth surface~\cite{bioucasdias2013hyperspectral}. It affords detailed physical spectral information in a narrow range of continuous wavelengths~\cite{xu2019abundance-indicated}. 
%This high spectral resolution can be exploited to develop discriminative measures to distinguish different materials~\cite{chang2007hyperspectral}. 
This kind of high spectral resolution can be employed to exploit differential measurements to distinguish between different materials.
The rapid development of HSI and processing techniques promotes the applications of remote sensing in national defense, disaster monitoring, precision agriculture and other fields~\cite{zhang2014hyperspectral}.

As a major application of HSI, image classification plays a significant role in hyperspectral data analysis~\cite{gu2017multiple}.
%Given a group of observations from an HSI, the goal of classification is to distribute a unique predefined class label to every pixel.
The goal of classification is to distribute a unique predefined class label to every pixel in a given group of observations from HSIs. 
%Especially for the precise classification of the surface of the Earth, the rich spectral information contained in HSI offers great advantages for material discrimination and thus facilitates good classification performance~\cite{goetz1985imaging}. 
Particularly for the accurate classification of the Earth's surface, the wealth of spectral information contained in HSI provides a great advantage for material identification, thus promoting good classification performance.
However, there exist two main challenges in HSI classification~\cite{gao2018tensorized}. 
%The first is the so-called curse of dimensionality. This effect is caused by the unbalance between the high dimensionality of the data and the small number of available labeled samples, which leads to ill-posed problems that are extremely formidable to tackle.
%~\cite{bioucasdias2013hyperspectral}. 
The first is the so-called curse of dimensionality. The imbalance between the high dimensionality of hyperspectral data and the few available labeled samples leads to this phenomenon. This results in ill-posed problems which are extremely formidable to tackle.
%The second is the characterization of mixed pixels, which are inherently contained in the data~\cite{imbiriba2016nonparametric}. 
The second is the mixed characteristics of pixels, which are included in the data inherently.
%As each pixel in the HSI is determined by the combination of different substances, it is difficult to categorize the samples in the original feature space~\cite{bioucasdias2012hyperspectral}. 
%Since each pixel in the HSI is determined by a combination of different substances, it is difficult to classify samples in the original feature space.
It is tough to classify samples in the primeval feature space, since the combination of different substances determines each pixel in HSI.
%These challenges in HSI classification call for the development of advanced methods~\cite{zhang2018exploiting}. 
These challenges in HSI classification demands the proposals of advanced methods.

In the preliminary stage of HSI classification, the main approaches center at exploring the function of spectral characteristics of HSI. Therefore, plentiful pixel-wise classification approaches (such as support vector machines (SVM)~\cite{archibald2007feature}, neural networks~\cite{zhong2012an} and dynamic or random subspace~\cite{du2014target}) are proposed. 
Besides, some other classification methods concentrated on designing a valid dimension reduction or feature extraction technique, such as principal component analysis (PCA)~\cite{licciardi2012linear}, and linear discriminant analysis (LDA)~\cite{bandos2009classification}. 
Since the spatial contexts are not taken into account, these pixel-wise classifiers yield unsatisfactory classification maps.
%Recently, spatial features have been reported to be very instrumental in improving the representation of hyperspectral data and increasing the classification accuracies~\cite{ghamisi2018new}. 
%Recent studies have indicated the utility of spatial features for improving the representation of hyperspectral data and increasing the classification accuracy~\cite{ghamisi2018new}.
Recent studies have indicated the utility of spatial features in improving hyperspectral data representation and raising classification accuracy~\cite{ghamisi2018new}.
%An increasing number of spectral-spatial feature-based classification (SSFC) frameworks have been developed to integrate spatial context information into pixel-wise classifiers.
In order to integrate spatial contextual information into pixel-level classifiers, an increasing number of classification frameworks based on spectral-spatial features have been developed.
%For instance, extended morphological profiles (EMPs) are used to exploit the spatial information via multiple morphological operations~\cite{li2013generalized}. 
For instance, the extended morphological profile  (EMP) employs multiple morphological operations to take advantage of the spatial information~\cite{li2013generalized}.
%The spatial information within a neighboring region was incorporated into the sparse representation model in~\cite{fang2017hyperspectral}.
In~\cite{fang2017hyperspectral}, the sparse representation model incorporates spatial information in neighboring regions.
%The basis of these sparse representations is the observation that hyperspectral pixels can usually be represented by a linear combination of several common pixels from the same class.
Hyperspectral pixels can be represented as linear combinations of ordinary pixels of the same category, which are the basis of these sparse representations.
%Furthermore, based on the similarity of intensity or texture, HSIs are divided into numerous superpixels to explore its spatial consistency~\cite{FangExtinction}. 
Furthermore, according to the similarity of intensity and texture, HSI is divided into multiple super pixels to explore its spatial consistency.

%Spectral-spatial classification methods can be classified into two categories. 
The classification methods based on spectral-spatial can be divided into two categories.
%The first category deploys the spectral and spatial information separately in which the spatial information is perceived in advance by the use of spatial filters~\cite{Ghamisi2015A}.
In the first classification method, spectral and spatial information are deployed respectively, in which spatial information is pre-perceived by employing spatial filters~\cite{Ghamisi2015A}.
%Then, these spatial features are added to the spectral data at each pixel.
Then, for each pixel, these spatial features are fused into spectral data.
Afterwards, the dimension reduction method may be employed prior to the ultimate classification.
%Spatial information can also be used to refine the initial pixel-level classification results as post-processing steps, for example, through mean shift~\cite{Deng2015Remote} or Markov random field (MRF)~\cite{5464269}, which is a typical strategy in image classification.
Spatial information can also be utilized as a post-processing step to improve the initial pixel-level classification results, for example, Markov random field (MRF)~\cite{5464269} and through mean shift~\cite{Deng2015Remote}, which are representative strategies in HSI classification.
%The second category combines spectral and spatial information for classification and segmentation.
The second method is to combine the spectral information and the spatial information to classify HSI.
%In~\cite{Li2012Spectral}, Li \emph{et al.} proposed to integrate the spectral and spatial information into a Bayesian framework, and then use either supervised or semi-supervised algorithms to perform an additional step to improve the initial classification results. 
In~\cite{Li2012Spectral}, firstly spatial information and spectral information are integrated into the Bayesian framework, and then supervised or semi-supervised algorithms are adopted to perform additional steps to optimize the primary classification results.
%Because hyperspectral data are normally represented by 3D cubes, the second method can produce a large number of features with discriminative information, which can effectively improve the classification performance.
Because 3D cubes are usually employed to represent hyperspectral data, the second method can produce numerous features with discriminative information, which is beneficial to improve the classification results of HSI.

Although most HSI classification methods based on spectral-spatial achieved superior performance, they are heavily dependent on handcrafted features or shallow descriptor. 
However, majority of the handcrafted features are typically designed for particular tasks and rely on expertise in the parameter setting stage, which inhibits the applicability of these methods in complicated scenarios. 
Besides, the expressiveness of handcrafted features may not be sufficient to distinguish the fine distinctions between different classes or the large differences between the same class. 
In HSI classification, extracting as many discriminant features as possible is the key requirement.
%Extracting more discriminant features is the key critical demand of HSI classification.

In recent years, deep learning (DL) has become a vigorous development trend in the field of big data analysis, and major breakthroughs have been made in lots of computer vision tasks ( such as object detection, image classification, and natural language processing). 
%~\cite{szegedy2015going}~\cite{girshick2014rich}~\cite{bordes2012joint}
Under the stimulation of these successful applications, DL is introduced into HSI classification and satisfactory results are obtained. 
Compared with conventional handcrafted strategies, DL technology can employ a series of hierarchical layers to obtain informative features from the initial data. 
Specifically, the simple features such as edge and texture information can be extracted by the early layer, while the more complex features can be handled by the deeper layer. 
The entire learning procedure is automatically driven, so that deep learning technology can better cope with various situations. 
%In general, deep learning is recognized as an effective feature extraction approach during HSI classification, while different networks focus on extracting different feature types~\cite{DBLP:journals/tgrs/LiSFCGB19}.
Generally speaking, in HSI classification, DL is consider as a valid feature extraction methods, and different feature types can be extracted by different networks specifically.

In the past decades, a multitude of DL-based HSI classification methods have been proposed. 
Among miscellaneous DL-based models, convolutional neural networks (CNNs)~\cite{Petersson2016Hyperspectral} are widely used in pixel level labeling. 
%The majority of the existing CNN-based HSI classification methods have been generalized to take both spectral and spatial information into considerations in a single classification framework. 
Most of the existing HSI classification approaches based on CNN consider the spectral and spatial information in a unitary classification framework
%In some HSI classification methods based on CNN~\cite{Wenzhi2016Spectral}, the spatial features are obtained by a 2D-CNN model that exploits the first few principal component bands of the original hyperspectral data. 
In some HSI classification methods based on CNN~\cite{Wenzhi2016Spectral}, the spatial features can be achieve via adopting the 2D-CNN model, in which the first several principal component bands of the primitive hyperspectral data can be exploited.
%Yu \emph{et al.} ~\cite{yu2017convolutional} proposed a CNN architecture that uses a convolutional kernel to extract spectral features along the spectral dimension only. 
A CNN architecture that employs convolutions kernels to obtain the spectral features only along the spectral dimension has been proposed by Yu \emph{et al.} in~\cite{yu2017convolutional}.
%To obtain features in the spatial domain, normalization layers and a global average pooling layer are applied. 
The global average pooling layers and the normalization layers are applied to obtain the feature in the spatial field. 
%Besides, a 3D-CNN can learn the signal changes in both spatial and spectral dimensions of local spectral images. 
Besides, the signal changes of local spectral images in the two dimensions of spectral and spatial can be learned by 3D-CNN.
%Therefore, it can extract significant discriminative information for classification and exploit powerful structural characteristics for hyperspectral data. 
Therefore, 3D-CNN can take advantage of the formidable structural features of hyperspectral data to extract important discriminative information and perform classification.
%In~\cite{chen2016deep}, Chen \emph{et al.} further proposed a 3D-CNN-based feature extraction model with combined regularization (3D-CNN-LR) to extract effective spectral-spatial features for HSI classification. 
In order to extract the spectral spatial features in hyperspectral data more effectively, Chen \emph{et al.} combined the regularization on the basis of 3D-CNN for HSI classification (3D-CNN-LR) in~\cite{chen2016deep}.
%However, the generalization ability of a relatively small network is usually lower than that of deep networks.  

Traditional CNN-based HSI classification methods often utilize a local patch for each pixel to execute the training and prediction steps, which may increase the calculation burden due to repeated computation.
Besides, the uncertainty size of the local patch could decrease the classification accuracy inevitably.
%Since CNN can efficiently discover the spatial structures between the neighboring patches of the input data, the resulting classification maps generally appear smoother in spite of not modeling the neighborhood dependencies directly~\cite{alam2019conditional}.
Since the spatial structure between adjacent patches of input data can be effectively observed by CNN, even if the neighborhood dependency is not directly modeled, the generated classification map will mostly appear smoother~\cite{alam2019conditional}.
%However, the possibility of reaching a local minimum during CNN training and the existence of noise in the input image may generate holes or isolated areas in the classification map.
However, the classification map may generate holes or isolated regions, due to the existence of noise in the input image and the probability of reaching a local minimum during the training of CNN.
%Compared with other machine learning methods, CNN-based methods are normally limited by the shape and edge constraints, which will trigger the situation that the final classification map is rough on the edge.
In addition, CNN-based methods are normally limited by the constraint of shape and edge, resulting in rough edges on the final classification map.
%Moreover, in HSIs, cloud shadows and topography cause variations in contrast, which frequently generate incorrect classification in images. 
Moreover, in HSIs, topography and cloud shadows lead to contrast changes, which can result in incorrect classifications.
%Due to these reasons, CNN sometimes recognizes only parts of the regions properly~\cite{Douglas2013A}.
Due to afore mentioned reasons, CNN occasionally can correctly recognize part of the areas only.

Therefore, despite increasing training data and more complicated network architectures being designed, there is a technical barrier in the application of CNN in HSI classification--contextual information. 
%It has been well recognized for years in computer vision domain that contextual information, or relation, is capable of offering important cues for HSI classification tasks.
Contextual information or relationships can provide important clues for HSI classification tasks, which has been widely recognized in the field of computer vision.
%Under these circumstances, spatial relations are of paramount importance, especially when a domain in the image demonstrates significant visual ambiguities. 
In this situation, the spatial relationship is crucial, especially when a certain domain in the image demonstrates obvious visual ambiguity.
%To address this issue, several attempts have been made to introduce spatial relations into networks by using either graphical models~\cite{alam2019conditional} or spatial propagation networks~\cite{pan2018spatial}.
In order to solve this problem, some researchers have tried to introduce spatial relationship into network by adopting spatial propagation network~\cite{pan2018spatial} or graphical model~\cite{alam2019conditional}.
%Nevertheless, these methods seek to obtain global spatial relations implicitly with chain propagation ways, whose effectiveness depends heavily on the learning effect of long-term memorization. 
Nevertheless, these approaches attempt to implicitly obtain the global spatial relationships through chain propagation, and the learning effect of long-term memory heavily determines its effectiveness.

To eliminate the existing drawbacks of the traditional CNN-based method, we propose a HSI classification approach based on a fully convolutional spatial propagation network (FCSPN). 
Through utilizing the 3D fully convolutional network (3D-FCN), the entire input image can be learned, and the learning process can be completed by feedforward calculation and back propagation simultaneously. In that case, not only it can alleviate the problems of repeated calculation and unfitness size of local patches caused by traditional local patch-based learning method, but also it can accelerate the learning speed. Moreover, a novel dual separable residual (DSR) unit can effectively extract feature information via considering both spectral and spatial information with fewer parameters, while the channel-wise attention mechanism can grasp the most informative channels from redundant channel information. 
%Furthermore, the spatial correlation of HSI is captured by the convolutional spatial propagation network (CSPN) to maintain the spatial consistency of HSI, which can further refine the classification accuracy. 
Furthermore, the convolutional spatial propagation network (CSPN) captures the spatial correlation of HSI, maintains the spatial consistency of HSI, and further refines the classification performance.

In summary, this paper has the following contributions:
\begin{itemize}
  \item [1)] 
  To the best of our knowledge, this is the first time to integrate 3D-FCN with CSPN for HSI classification. On the basis of this method, the spatial consistency in HSI can be fully utilized to improve classification accuracy, while avoiding generous redundancy in computation and storage.
  \item [2)]
  With the proposed 3D-FCN, HSI is trained end-to-end and pixel-to-pixel, which effectively reduce computational complexity. Besides, the arbitrary-sized image can be imported into the framework without considering the size of the local patch. Consequently, it will enhance the adaptive capacity of the proposed method.
  \item [3)]
  In the proposed 3D-FCN, a novel dual separable residual (DSR) unit is introduced to effectively extract feature information via considering both spectral and spatial information with fewer parameters, meanwhile, the channel-wise attention mechanism is deployed to emphasize useful channel information and suppress less useful channel information.
  \item [4)]
  The CSPN is adopted to capture the spatial correlations of HSIs so that the HSI spatial consistency can be maintained, thus the classification results can be further refined.
\end{itemize}

In the following, the proposed FCSPN method is described detailedly in Section \uppercase\expandafter{\romannumeral2}. Then experimental results are presented for performance evaluation on three HSI benchmark datasets in Section \uppercase\expandafter{\romannumeral3}. Finally, the conclusion is provided in Section \uppercase\expandafter{\romannumeral4}.

\section{Proposed Method}
\subsection{Overview of the FCSPN Architecture}
\begin{figure*}[htp]
 \centering
 \includegraphics[scale=1.13]{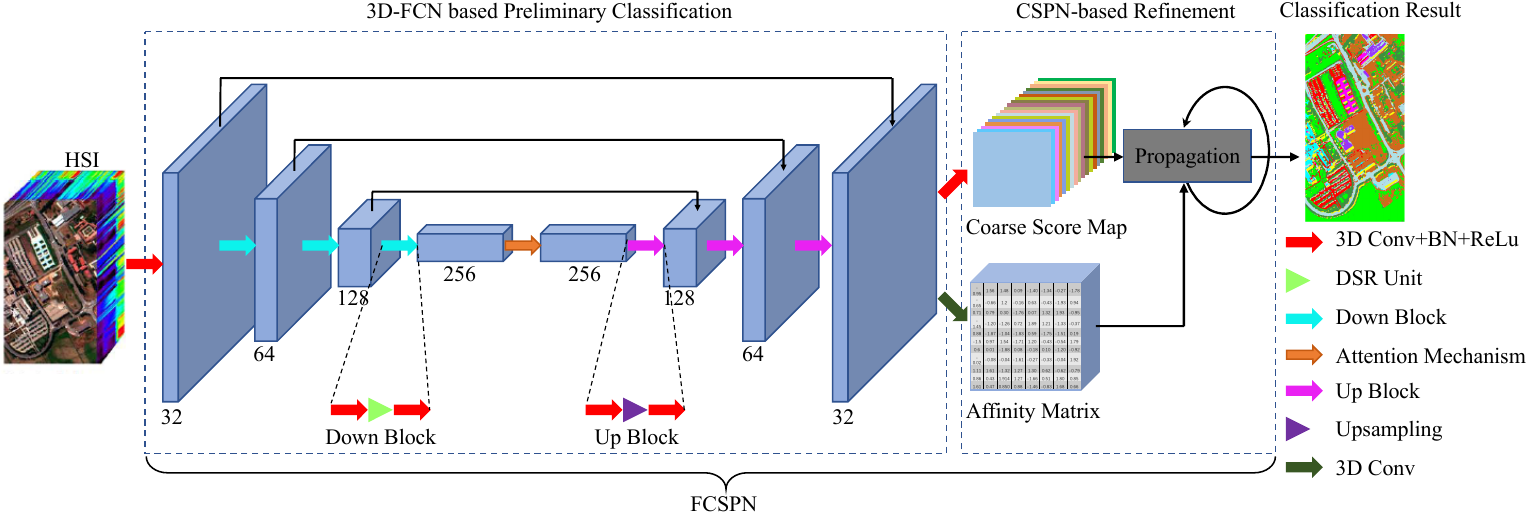}
 \vspace{4pt}
 \setlength{\abovecaptionskip}{-10pt}
 \caption{FCSPN based HSI classification framework. There are two main parts in this framework, which are the 3D-FCN based preliminary classification, and the CSPN-based refinement.}
 \label{patch_size}
\vspace{-0.55cm}
\end{figure*}
The proposed HSI classification framework has two major parts as shown in Fig.1.
One is the 3D-FCN based preliminary classification, and the other is the CSPN-based refinement.

%In the first part, the 3D structure of HSI is well described based on 3D-FCN, and the robust global feature can be extracted with DSR unit and the spectral-wise attention mechanism learning strategy so that a coarse pre-classification result can be obtained.
For the 3D-FCN based preliminary classification, inspired by successes of FCN~\cite{Long2015Fully} architecture, we adjust the architecture and extend it to the 3D-FCN, which can learn more efficiently from whole HSI inputs. Based on the 3D-FCN, the 3D structure of HSI is well described, in which both spectral and spatial feature information can be processed by DSR unit with fewer parameters, so the most informative channels can be extracted with the channel-wise attention mechanism, thus a coarse pre-classification result can be obtained.

For the CSPN-based refinement, the refinement method based on CSPN is employed to heighten the classification accuracy. 
%It is well known that HSI classification is based on pixel. Nevertheless, the spatial information in HSI is proved to be effective in improving the classification accuracy~\cite{LiSpectral,YueSpectral}. 
Since the first part of preliminary classification based on the 3D-FCN, the spatial consistency in HSIs is not fully taken into account, while the spatial information is profitable to increase the classification result.  
To address this shortfall, we adopt CSPN to describe spatial consistency in HSI, and the spatial consistency can be maintained; and the classification results can be further refined.

\subsection{3D-FCN based preliminary classification}
Due to that the structure of HSI is 3D, the 3D classification model of HSI can be realized intuitively. 
%Fortunately, in the 3D-FCN framework, each layer of data in a convnet is a 3D array of size $ m \times n \times d $, where \emph{m} and \emph{n} are spatial dimensions, and \emph{d} is the feature or channel dimension. The first layer is the image, with pixel size \emph{m} and \emph{n}, and \emph{d} color channels. 
Fortunately, in the 3D-FCN framework, in the convolutional network each layer of data is a 3D array represented by $ m \times n \times d $, where the spatial size is denoted by \emph{m} and \emph{n}, and the feature or channel size is indicated by \emph{d}. The first layer is an image with pixel sizes of  \emph{m} and \emph{n}, and the color channel is \emph{d} .
It is reasonably assumed that there is a natural correlation between the HSI and the 3D-FCN; specifically, the spatial dimension and feature or channel dimension in the 3D-FCN correspond to the spatial and spectral dimensions in HSI respectively. Therefore, an entire HSI can be input into a network framework by the 3D-FCN for global feature extraction to accelerate the speed of model training.
%then, a coarse score map of $ m \times n \times c $-size can be achieved, where \emph{c} represents the number of class of HSI.
%The 3D-FCN based preliminary classification strategy enables the whole original HSI to be the input data, which has the ability to accelerate the efficiency of model training. 

\begin{figure}[htp]
 \centering
 \includegraphics[scale=0.8]{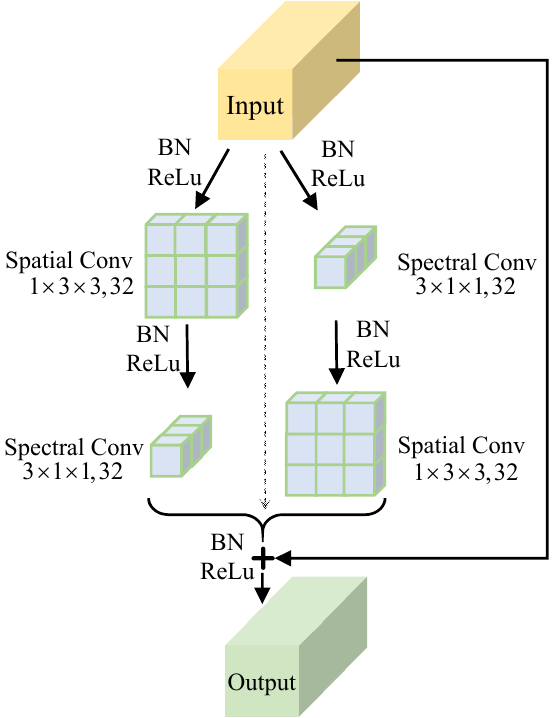}
 \vspace{0.25cm}
 \setlength{\abovecaptionskip}{-7.5pt}
 \setlength{\belowcaptionskip}{2.5cm}  
 \caption{The architecture of the DSR unit.}
 \label{patch_size}
\vspace{-0.25cm}
\end{figure}

A detailed illustration of the 3D-FCN based preliminary classification can be found in Fig.1. 
Firstly, the entire HSI image is taken as input, features can be extracted through a 3D convolution layer, the normalization is operated by the batch normalization (BN) layer and the linear operation is handled by the rectified linear unit (ReLu)~\cite{krizhevsky2012imagenet} activation layer successively. The convolution kernel in 3D convolution layer with the size of $ 5 \times1\times1$, and the stride is $(5,1,1)$.
Then followed by three Down Blocks, the detailed construction of each Down Block is demonstrated in Fig.1.

It is worth mentioning that in each Down Block, the convolution kernel in the prior 3D convolution layer is $ 3 \times3\times3$, and the stride is $(2,1,1)$, while in the latter 3D convolution layer, the convolution kernel is $ 1 \times3\times3$ with the stride is $(1,2,2)$.
Based on these convolution operations, the number of channels in the feature map will double with each Down Block.

\noindent{\bf DSR Unit} In spired by the outstanding performance of the residual network (ResNet)~\cite{he2016deep}, we adopt the basic structure of the ResNet while designing this strategy. 
Regarding the 3D characteristics of the HSI, 2D-ResNet is extended to 3D-ResNet for HSI classification~\cite{jiang2019hyperspectral}.  A series of dual separable residual (DSR) units are adopted to expose the spatial-spectral characteristics of HSI explicitly. Fig.2 displays the structure of the DSR unit, which includes two main branches. 
In the left branch, the spatial information is extracted first followed the spectral information, specifically, the left branch contains a 3D spatial convolution layer with the kernel size is $ 1 \times 3 \times 3 $, while a 3D spectral convolution layer with $ 3 \times1\times1 $ kernel size. Correspondingly, both the spectral and spatial convolution steps are executed in the first and second order in the right branch, the sizes of convolution kernels are the same as the left branch. Each convolution layer is followed by the BN layer and the ReLu active layer successively. Under this structure, the spatial-spectral characteristics of HSI can be well extracted with fewer parameters. For instance, in the standard 3D-ResNet, for a 3D convolution kernel with the size of $ M \times M \times M $, the parameter quantity is $ M ^{3}$. By comparison, in our proposed DSR unit, the quantity of parameter is only $ (M ^{2} + M) \times 2$, thus it can be seen that the DSR unit cuts down the computational cost by reducing the number of pamameters. 

\begin{figure}[htp]
 \centering
 \includegraphics[scale=0.8]{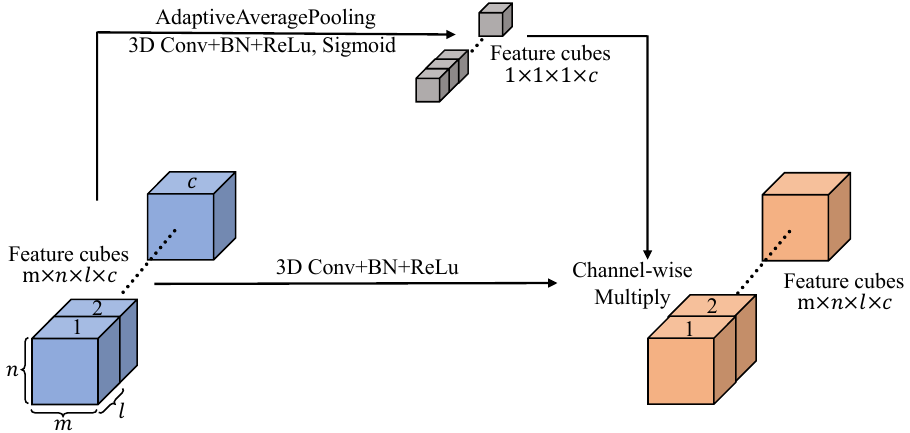}
 \vspace{0.25cm}
 \setlength{\abovecaptionskip}{-7.5pt}
 \setlength{\belowcaptionskip}{2.5cm}  
 \caption{The basic structure of channel-wise attention mechanism.}
 \label{patch_size}
\vspace{-0.25cm}
\end{figure}
%After a series of feature extraction based on DSR unit, we can get a 3D feature map. 
With 3D convolutional network based HSI classification, hundreds of channels are straight employed as the input data for convolutions, which will necessarily carry redundant information between channels.
To alleviate this issue, we adopt the channel-wise attention mechanism to obtain the most informative channels.
Fig.3 illustrates the basic structure of channel-wise attention mechanism.
The feature cubes pass through the 3D convolutional layer with $ 3 \times3\times3$-sized kernel, going through the 3D Adaptive average pooling layer with the output size is $ 1 \times1\times1$, the convolution kernel size of the 3D convolution layer is $ 1 \times1\times1$, BN layer, ReLu activation layer and Sigmoid functions successively.
%Finally, do the multiplication on the channel-wise.
At the final step, a multiplication is processed on the channel-wise.
Consequently, the channels with more abundant information will gain higher weight.
%The objective of this mechanism is to selectively emphasize informative channels and suppress less useful channels. 
The objective of this mechanism is that to emphasize channels with more information and suppress channels with less information.

%After the high-dimensional feature maps are obtained, the upsampling modules use trilinear interpolation to upsample the feature maps to the same size as the original image. However, spatial information is weaken or lost with the DSR unit during the forward process of the ResNet in~\cite{he2016deep}. Thus, we add mirror connections similar with the U-shape network~\cite{cicek20163d} by directed concatenating the feature from the DSR unit to the upsampling module.
After the high-dimensional feature maps are achieved, the Up Blocks are used to upsample the feature maps to get the same size as the original image.
%Specifically, each Up Block is successively composed of Pub Unit, upsampling module and Pub Unit. 
Fig.1 shows the detailed structure of the Up Block.
Corresponding to the Down Block above, in the prior 3D convolution layer of the Up Block, the kernel size of 3D convolution is $ 5 \times1\times1$ with the stride of $(2,1,1)$, while in the latter 3D convolution layer, the kernel size is $ 3 \times3\times3$ with the stride of $(1,1,1)$.
%The upsampling modules use trilinear interpolation.
Notably, we used trilinear interpolation in the upsampling module.
%However, spatial information is weaken or lost with the DSR unit during the forward process. 
%It is considered that the spatial information is weaken or lost with the DSR unit during the forward progress. 
%Furthermore, we add mirror connections similar to the U-shape network~\cite{cicek20163d} by directed concatenating the feature from the Down Block to the Up Block.
Furthermore, by concatenating the features between the Down Block and the Up Block directly, mirror connections similar to the U-shape network~\cite{cicek20163d} are added. 
After passing through three Up Blocks, the size of the gained feature map is the same as the input image, so that a prediction can be generated for each pixel.

\subsection{CSPN-based refinement}

After the preliminary classification based on the 3D-FCN, a coarse classification result can be achieved.
As 3D-FCN pays more attention to the global feature of the HSI, detailed local spatial characteristic in HSI has not been well extracted in the above section which results in an inappropriate classification result. To provide a robust way to describe the spatial consistency feature within the HSI, the CSPN based refinement strategy is proposed. 

Learning spatial propagation with networks has attracted great interests in recent years~\cite{pan2018spatial}. 
%Maire \emph{et al.}~\cite{7780395} tried to predict entities of an affinity matrix directly by learning a CNN, which presents a good performance. 
Maire \emph{et al.}~\cite{7780395} attempted to directly predict the entities in the affinity matrix by learning CNNs, and presented good achievement.
%However, since the affinity is followed by a non-differentiable, independent solver of spectral embedding, it cannot be used for the end-to-end classification tasks.
However, the affinity can not be deployed for the end-to-end classification because it is followed by the non-differentiable independent solver of the spectral embedding.
%Liu \emph{et al.} examined a CNN model to learn a task-dependent affinity matrix by converting the modeling of affinity to learning a local linear spatial propagation, yielding a simple, yet effective approach for the enhancement of results~\cite{liu2017learning}. 
Based on a CNN model, Liu \emph{et al.} studied to learn a task-related affinity matrix by transforming the affinity model into learning the local linear space propagation, thereby deriving a simple and effective method to enhance the performance~\cite{liu2017learning}.
%However, their affinity matrix requires additional supervision of ground-truth sparse pixel pairs, which limits the potential connections between pixels.
However, the potential connections between pixels are limited, because the affinity matrix requires extra monitoring of sparse pixel pairs on the ground-truth .

%Inspired by these advanced methods, the CSPN-based refinement strategy is adopted to capture the spatial consistency in HSI, and the spatial consistency can be maintained; then the classification results can be further increased.
Inspired by these advanced methods, a refinement strategy based on CSPN is employed to capture the spatial consistency in HSI, thereby maintaining spatial consistency and further improving the classification results.
%Specifically, an affinity matrix is employed to extract the consistent spatial features on each classification score map through a propagation way.  Additionally, we use an efficient linear propagation model, where the propagation is performed with a manner of recurrent convolutional operation. Then, the refined score map can be achieved to get a more accurate classification result. 
Specifically, we employ an affinity matrix to extract the consistent spatial features on each classification score map through a propagation way. 
%Simultaneously, an effective linear propagation model is used, in which the propagation is performed with a manner of recurrent convolutional operation. 
Simultaneously, the recurrent convolutional operation is employed to perform effective the linear propagation model.
%Then, it enables the score map to be refined to obtain a more accurate classification result.
Then, it can refine the score map to obtain more accurate classification results.
%The affinity matrix is learned in the following way. 

%\noindent{\bf Affinity matrix} Many research studies have been carried out on learning affinity matrix with deep CNN for diffusion or spatial propagation due to its theoretical supports and guarantees~\cite{weickert1996theoretical}. 
\noindent{\bf Affinity matrix} Due to its theoretical support and guarantee, many researchers have carried out a lot of studies on diffusion or spatial transmission adopting deep CNN to learn affinity matrix~\cite{weickert1996theoretical}.
%The most related work with our approach is~\cite{cheng2018depth}, where the learning of a large affinity matrix for diffusion is converted to learning a local linear spatial propagation, yielding a simple while effective approach for classification improvement.
%The most relevant work for our method is~\cite{cheng2018depth}, where the learning of a large affinity matrix for diffusion is converted to learning a local linear spatial propagation, resulting in a simple and effective classification improvement approach. 
The most relevant work with our method is~\cite{cheng2018depth}, in which the learning of the large-scale affinity matrix of diffusion is transformed into the learning of the local linear spatial propagation, thus an effective and simple classification enhancement approach can be obtained.
%Specifically, the score maps at all pixels are updated simultaneously in the local convolution context, and the long-range context is achieved through a recurrent operation.
Specifically, in the local convolutional contextual the score map of all pixels can be update simultaneously, while a recurrent operation is deployed to achieve the long-range contextual.
\begin{figure}[htp]
 \centering
 \includegraphics[scale=0.54]{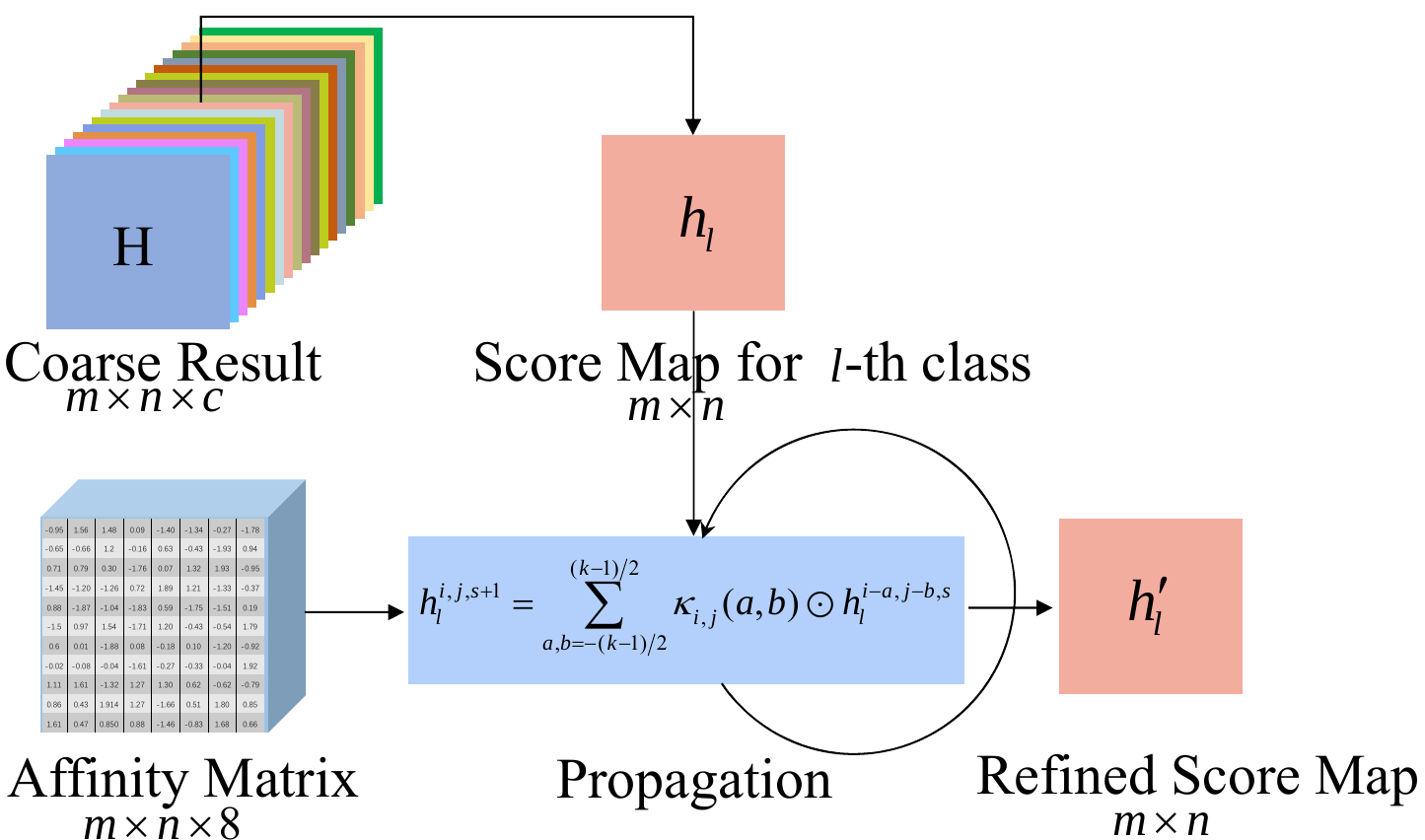}
 \vspace{4pt}
 \setlength{\abovecaptionskip}{-3pt}
 \caption{The propagation process of CSPN-based refinement.}
 \label{patch_size}
\end{figure}

Fig. 4 shows the propagation process of CSPN-based refinement. For a clear explanation, assume we obtained a score map of $l$-th class $h_{l}$ $\in \mathbb{R}^{m \times n}$, which has $c$ classes. 
The convolutional transform function for each spectral step $s$ with kernel size is $k$ can be written as, 

$h_{l}^{i, j, s+1}=\sum_{a, b=-(k-1) / 2}^{(k-1) / 2} \kappa_{i, j}(a, b) \odot h_{l}^{i-a, j-b, s}$

$\text { where, } \quad \kappa_{i, j}(a, b)=\frac{\hat{\kappa}_{i, j}(a, b)}{\sum_{a, b, a, b \neq 0}\left|\hat{\kappa}_{i, j}(a, b)\right|}\text { , }$

\begin{equation}
\kappa_{i, j}(0,0)=1-\sum_{a, b, a, b \neq 0} \kappa_{i, j}(a, b)
\end{equation}
where the transformation kernel $\hat{\kappa}_{i, j} \in \mathbb{R}^{m \times n \times 8}$ represents the output from the affinity matrix, where depends on the input image spatially. The element-wise product indicated as $\odot$. 
%According to~\cite{liu2017learning}, we normalize kernel weights between range of $(-1, 1)$ so that the model can be stabilized and converged by satisfying the condition $\sum_{a, b, a, b \neq 0}\left|\kappa_{i, j}(a, b)\right| \leq 1$. Finally, a stationary distribution is obtained by iteration.
According to~\cite{liu2017learning}, the weight of the kernel is normalized to the range of $(-1, 1)$, thereby the condition $\sum_{a, b, a, b \neq 0}\left|\kappa_{i, j}(a, b)\right| \leq 1$ can be satisfied to make the model stable and convergent.
Finally, a stationary distribution is obtained by iteration.
Additionally, CSPN adopts $k \times k$-sized local contexts to propagate local areas in all directions in each step simultaneously, and a larger range of contextual can be observed when performing the recurrent processing.
Therefore, the eight-neighborhood spatial information of each pixel can be well utilized to facilitate the classification accuracy. 

\section{Experiments}
\subsection{Data Description}
In this paper, to assess the performance of our proposed approach three HSI classification benchmark datasets were employed.
\begin{figure}[htp]
	\centering
	\setlength{\abovecaptionskip}{0pt}
	\subfigure[]{\includegraphics[width=2.8cm,height = 2.8cm]{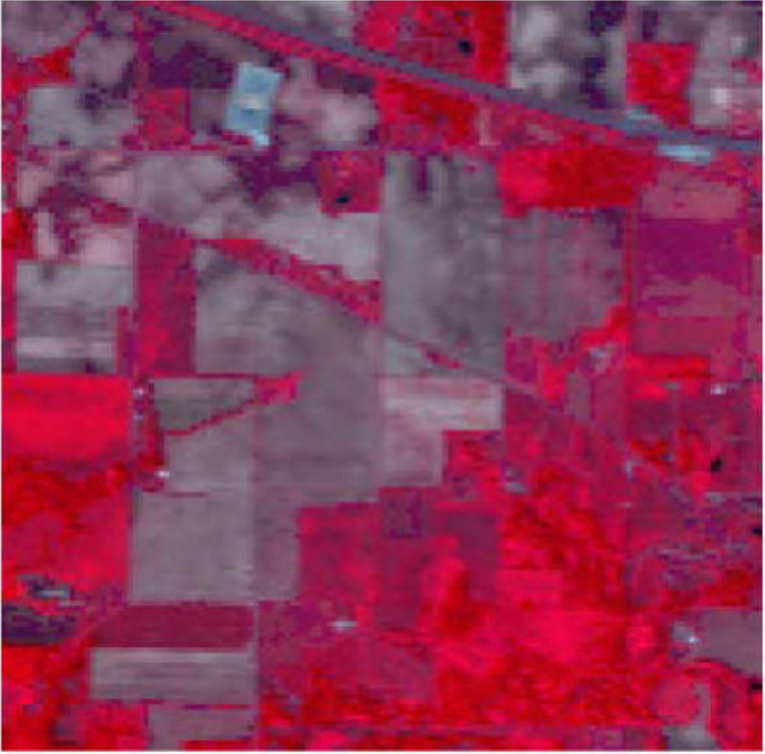}}
	\subfigure[]{\includegraphics[width=2.8cm,height = 2.8cm]{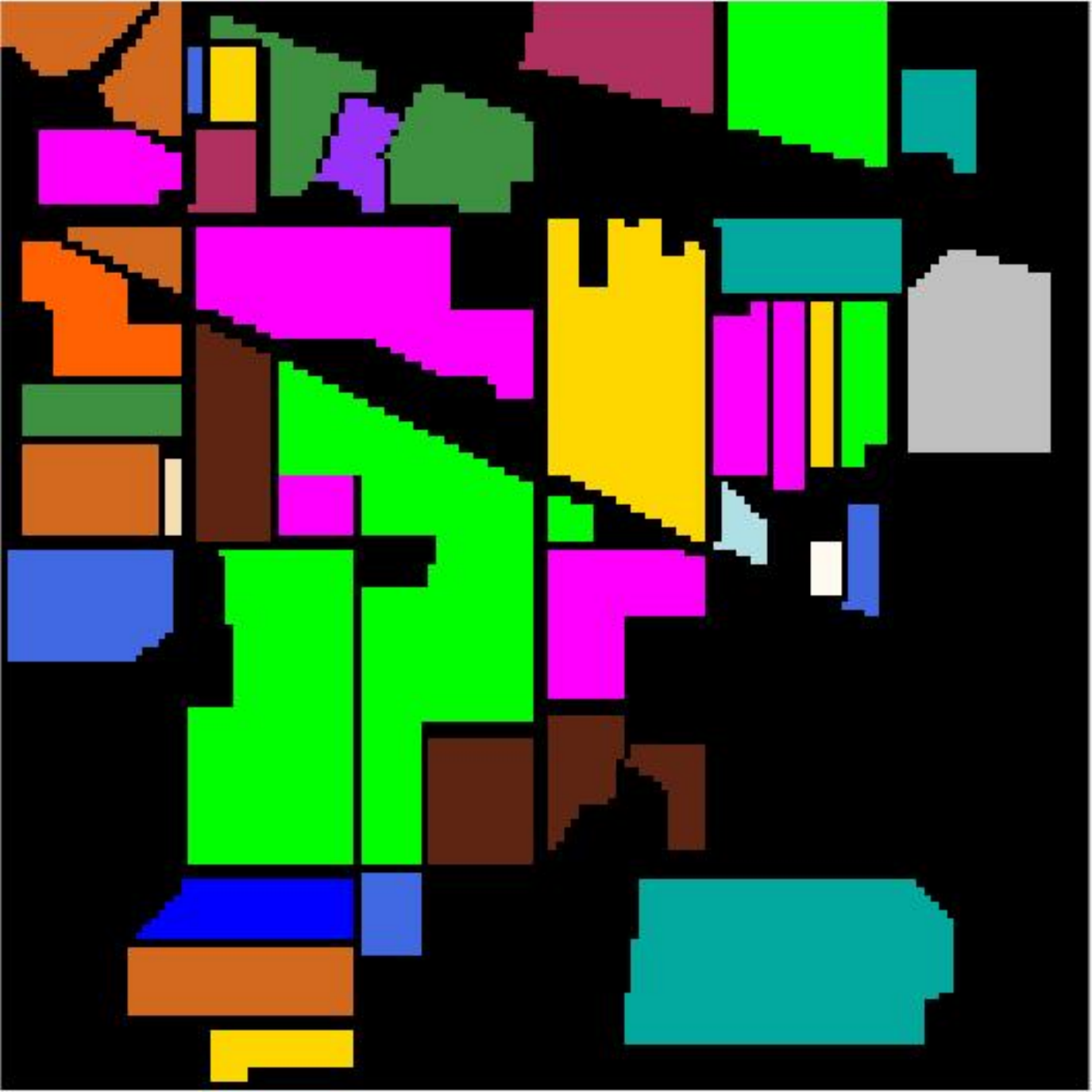}}
	\subfigure{\includegraphics[width=2.8cm,height = 2.8cm]{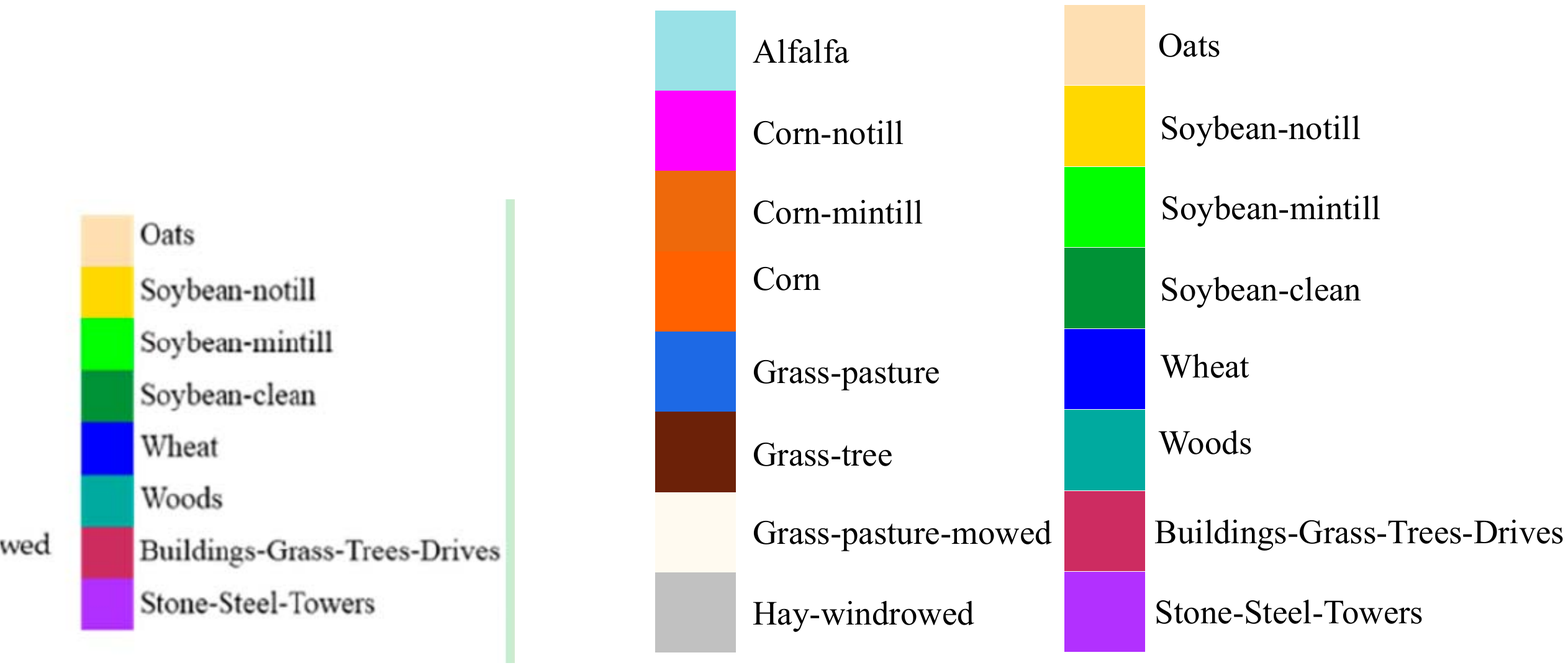}}
	\caption{Classification maps of Indian Pines dataset. (a) False color image. (b) Reference image.}
	\label{fig:camoflage_detection}
\end{figure}

The first dataset is Indian Pines, which was acquired by the Airborne Visible/Infrared Imaging Spectrometer (AVIRIS) sensor. The spatial size of this dataset is 145 pixels $\times$ 145 pixels, and there are 224 spectral channels with a wavelength range of 0.4–2.5 $\mu m$. By removing the water and noise absorption bands, the number of channels is reduced to 204. The ground truth provides 16 different classes of land cover. Fig. 5 shows The false color composite of the Indian Pines dataset and the corresponding ground reference data, respectively.

The second dataset is Salinas, which was collected over Salinas Valley by the AVIRIS instrument. The image is composed of 512 pixels $\times$ 217 pixels and 224 spectral bands, ranging from 0.4 to 2.5 $\mu m$. After the noisy bands and water absorption bands were removed, 204 bands were retained in the experiment. Sixteen classes were contained in the ground truth. The false color composite of Salinas image and the corresponding ground reference data are demonstrated in Fig. 6.

\begin{figure}[htp]
	\centering
	\setlength{\abovecaptionskip}{0pt}
	\subfigure[]{\includegraphics[width=3.1cm, height = 3.1cm]{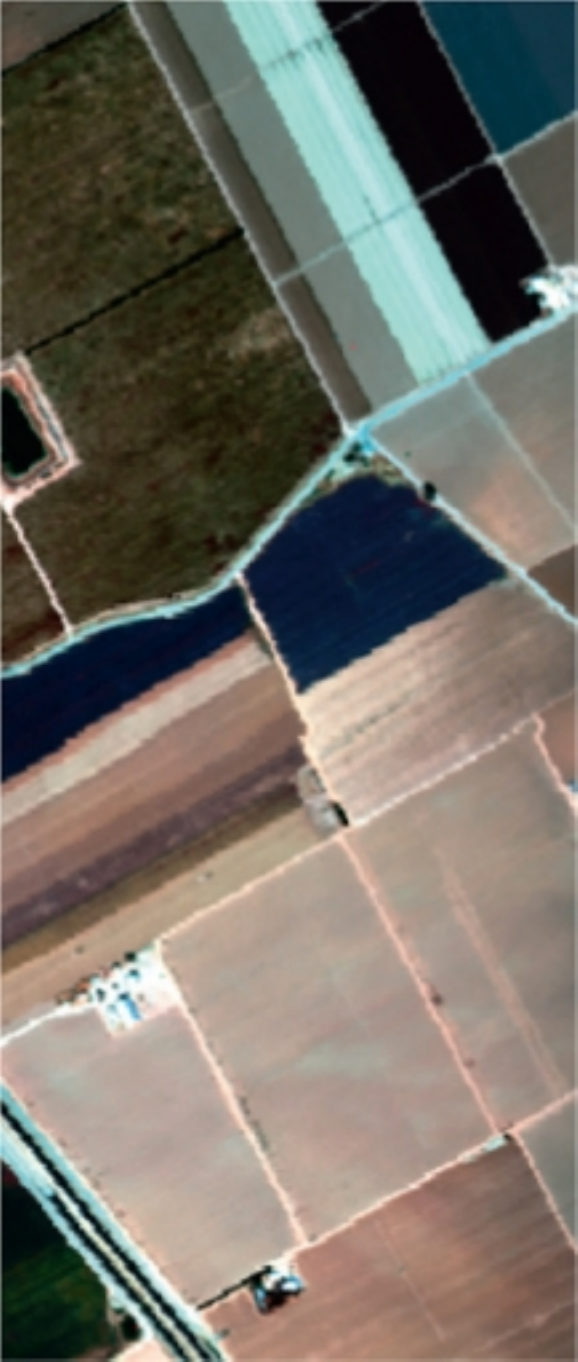}}
	\subfigure[]{\includegraphics[width=3.1cm, height = 3.1cm]{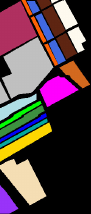}}
	\subfigure{\includegraphics[scale=0.25]{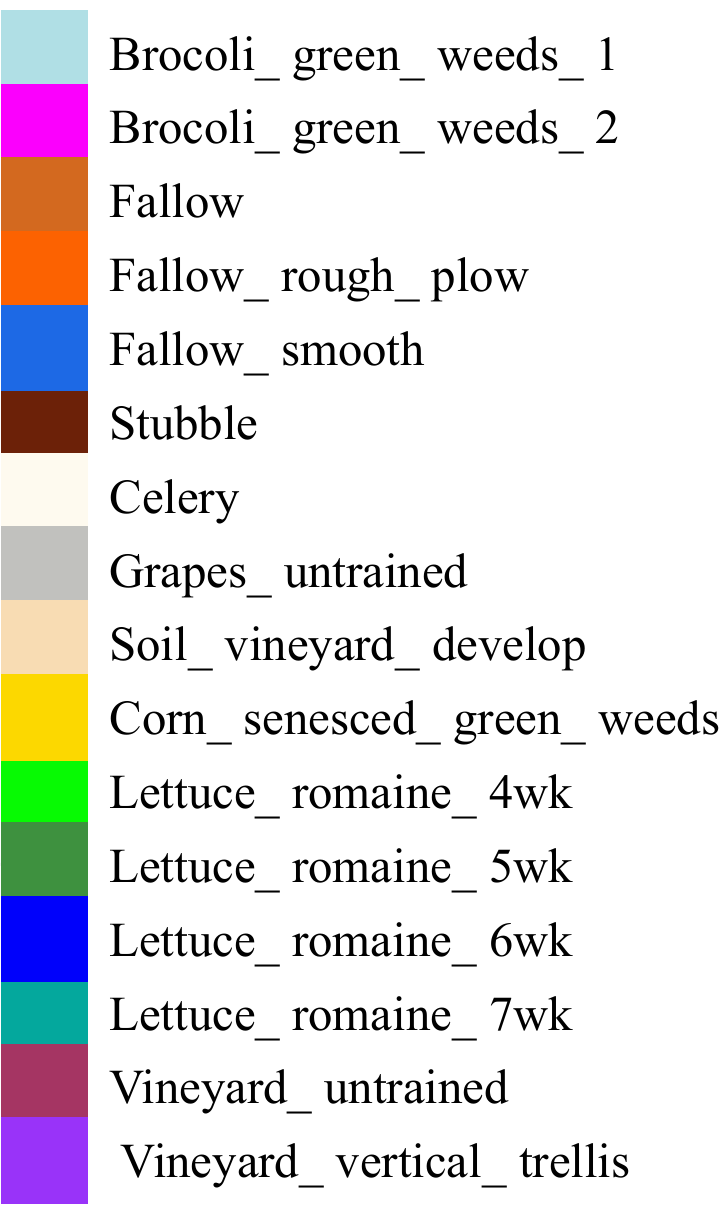}}
	\caption{Classification maps of Salinas dataset. (a) False color image. (b) Reference image.}
	\label{fig:camoflage_detection}
\end{figure} 
\begin{figure*}[htp]
	\centering
	\setlength{\abovecaptionskip}{0pt}
	\subfigure{\includegraphics[scale=0.7]{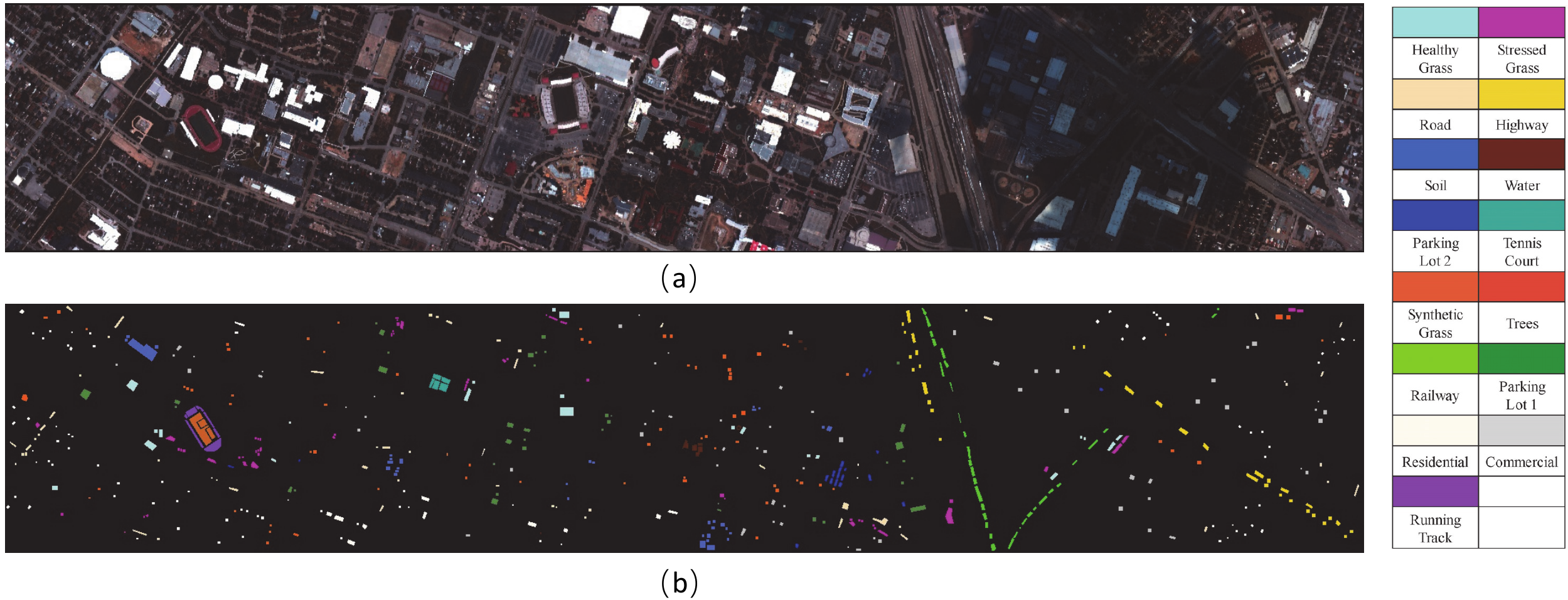}}
	\caption{Classification maps of University of Houston. (a) True color image. (b) Reference image.}
	\label{fig:camoflage_detection}
\end{figure*}

\begin{table*}[htp]	
\vspace{-0.25cm}
	\setlength{\abovecaptionskip}{0pt}
	\caption{Network architecture details of proposed FCSPN for Indian Pines Dataset (200 samples per class)}
	
	\centering
	
	\scalebox{1.22}{
		
		\begin{tabular}{c|c|c|c|c|c|c}
			
			\hline
			
			\multirow{2}{*}{Class} & \multicolumn{6}{c}{Methods}                                                                                                                                              \\ \cline{2-7} 
			
			& \multicolumn{1}{c|}{CCF~\cite{xia2017hyperspectral}} & \multicolumn{1}{c|}{3D-CNN~\cite{Li2017Spectral}}  &\multicolumn{1}{c|}{SSRN~\cite{zhong2017spectral}}  & \multicolumn{1}{c|}{LDCR~\cite{zhang2019learning}} & \multicolumn{1}{c|}{MSDN-SA~\cite{fang2019hyperspectral}}  & \multicolumn{1}{c}{FCSPN} \\ \hline
			
			Alfalfa     &      81.97 &  83.70  &    100     &84.20   &     100   &         100                  \\ 
			
			Corn-no till   &      82.49  &  73.06   & 95.11  &94.22 &  95.78&   99.86             \\ 
			
			Corn-min till  &    80.36 &  93.01  &  97.36    &91.75&    99.20 &  99.87              \\ 
			
			Corn   &    85.32  &   98.82 &  94.87  &100&   100    &100              \\ 
			
			Grass/pasture  &    84.89  & 99.75 &      96.92     &94.67&   98.66&       100              \\ 
			
			Grass/Trees  &    90.33   & 76.49 &    99.81       &98.87 &    100&  100             \\ 
			
			Grass-pasture-mowed &  84.14  &     93.92 &   100      &95.64 &   95.71  &  100               \\ 
			
			Hay-windrowed &    87.87 &     76.19  &    100  &99.64 &    100 &  100              \\ 
			
			Oats  &    84.43 &    99.33 &      100    &94.32  &  100  & 100             \\ 
			
			Soybean-no till &   79.32 &    86.42   &   99.10  &95.34  &   98.94  &   98.87                \\ 
			
			Soybean-min till  &   75.54 &    89.65&    99.95        &88.25  &    94.55 &   99.10             \\ 
			
			Soybean-clean till &  79.83 &    83.94  &    94.93      &96.95 &   99.39   &  99.33             \\ 
			
			Wheat & 87.93 &    94.97 &        100       &100  &   100 & 100              \\ 
		
			Woods &    85.46 &     95.49 & 99.81   &95.96	 &  99.23&     100              \\ 
			
			Buildings-grass  &   79.86   &       86.29 &        99.47           &93.01       &     99.89   &100                \\ 
			
			Stone-steel-towers &   89.42  &   90.68 &     100   &90.21      &   99.57          &100              \\   \hline
			
			OA(\%)  &   82.64  & 87.87&   98.41  &93.36  &  97.38         &    {\bf99.61}              \\ 
			
			AA(\%) &  83.27  &  86.32  &     98.58        &92.71  &   98.81        &{\bf 99.81}            \\ 
			
			$\kappa \times 100$   &   82.23&     83.86 &  98.13         &92.16 &   96.91     &   {\bf 99.56}               \\ \hline
			
		\end{tabular}
		
	}
	
	\label{In_200}	
\end{table*}

\begin{table*}[htp]	
	\setlength{\abovecaptionskip}{0pt}
	
	\caption{Network architecture details of proposed FCSPN for Indian Pines Dataset (5 percent of samples)}
	
	\centering
	
	\scalebox{1.22}{
		
		\begin{tabular}{c|c|c|c|c|c|c}
			
			\hline
			
			\multirow{2}{*}{Class} & \multicolumn{6}{c}{Methods}                                                                                                                                              \\ \cline{2-7} 
			
			& \multicolumn{1}{c|}{CCF~\cite{xia2017hyperspectral}} & \multicolumn{1}{c|}{3D-CNN~\cite{Li2017Spectral}} & \multicolumn{1}{c|}{SSRN~\cite{zhong2017spectral}} & \multicolumn{1}{c|}{LDCR~\cite{zhang2019learning}} & \multicolumn{1}{c|}{MSDN-SA~\cite{fang2019hyperspectral}} & \multicolumn{1}{c}{FCSPN} \\ \hline
			
			Alfalfa                      &      93.57                    &          90.21                      & 100       &        69.57                   &     94.69                                         &         97.83                   \\ 
			
			Corn-no till                      &      67.45                    &         70.42                  &  89.40           &         82.84                 &  91.12                                          &               96.92             \\ 
			
			Corn-min till                      &    67.54                      &      65.38                    &   96.75       &         70.84                  &   93.46                             &              99.76              \\ 
			
			Corn                      &    89.97                      &     87.69                       &  98.56          &       70.89                 &   92.07                                  &              100              \\ 
			
			Grass/pasture                      &    87.68                      &     86.47                      &    98.46         &         88.41               &   96.98                                       &              98.76              \\ 
			
			Grass/Trees                      &    93.28                      &     92.54                       & 92.27             &         95.34               &    99.98                                       &              99.04              \\ 
			
			Grass-pasture-mowed                      &  86.43                        &     83.86                        & 94.74            &         85.71                &   95.68                                  &             96.43               \\ 
			
			Hay-windrowed                      &    93.27                      &     86.43                        &  99.56                &         96.03            &    99.89                            &              100              \\ 
			
			Oats                      &    88.43                      &    78.79                       &   84.52          &         95.00                &  85.96                             &              95.00              \\ 
			
			Soybean-no till                     &   81.16                       &    87.24                       &  97.86          &         84.98                 &   93.73                             &            98.25                \\ 
			
			Soybean-min till                     &   60.78                       &    90.73                    &   91.28       &         80.41                   &    96.46                                   &              98.53              \\ 
			
			Soybean-clean till                    &  81.47                        &    85.97                       &   99.08      &         77.40                   &   93.46                                   &             94.10               \\ 
			
			Wheat                     & 94.53                         &    94.38                      &   100            &         99.51              &   99.23                          &              99.51              \\ 
		
			Woods                     &    87.62                      &     96.73                     &  99.42                      &         97.31      &  99.60                       &                   98.81               \\ 
			
			Buildings-grass                     &   76.74                       &        87.34                &     97.39             &         59.07           &     92.59                          &           96.90                 \\ 
			
			Stone-steel-towers                     &   93.24                       &    89.72                       &   92.55         &         83.87                  &    99.43                     &              95.70              \\ \hline
			
			OA(\%)                 &   78.69                       &  78.46                         &   93.97           &         83.87                &  95.68                          &              {\bf98.22}              \\ 
			
			AA(\%)                 &  84.46                        &  86.32                      &    90.46         &         83.57                &   95.27                          &               {\bf 97.84}            \\ 
			
			$\kappa \times 100$                  &   78.94                       &     73.86                        &  94.74           &         81.61               &   95.07                &            {\bf 98.22}               \\ \hline
			
		\end{tabular}
		
	}
	
	\label{In_005}	
\end{table*}

\begin{figure}[htp]
	\centering
	\setlength{\abovecaptionskip}{0pt}
	\subfigure[]{\includegraphics[width=2.8cm,height = 2.8cm]{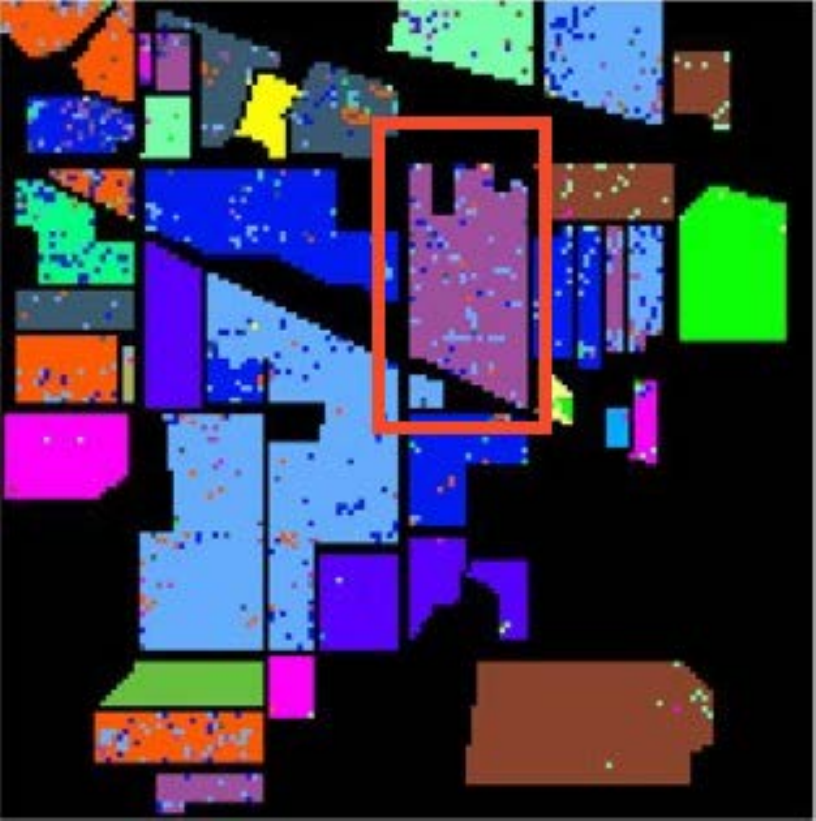}}
	\subfigure[]{\includegraphics[width=2.8cm,height = 2.8cm]{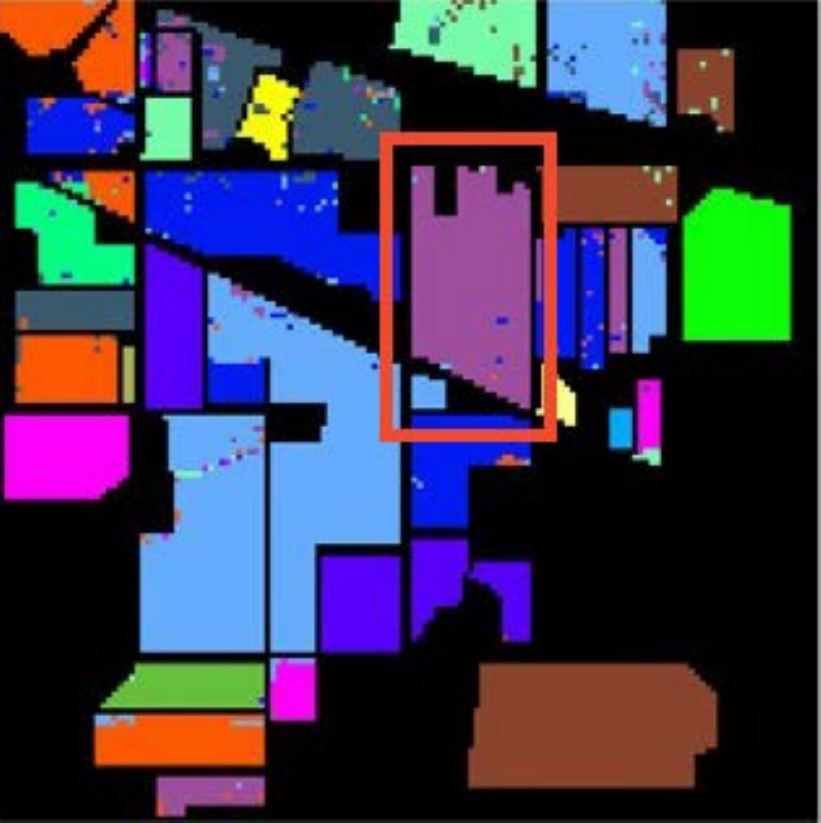}}
	\subfigure[]{\includegraphics[width=2.8cm,height = 2.8cm]{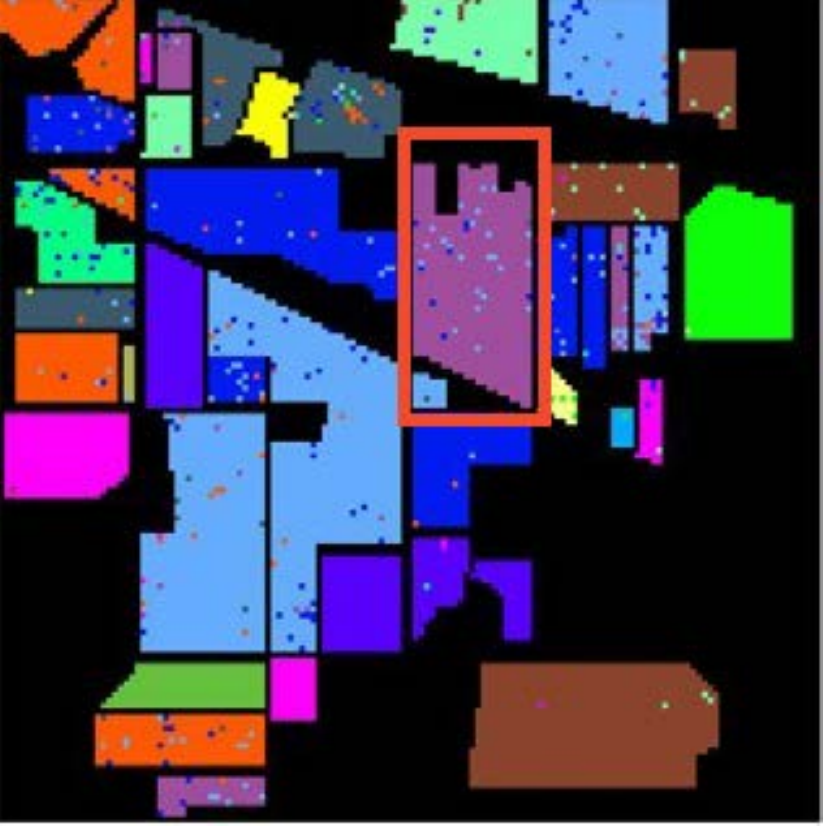}}
	\subfigure[]{\includegraphics[width=2.8cm,height = 2.8cm]{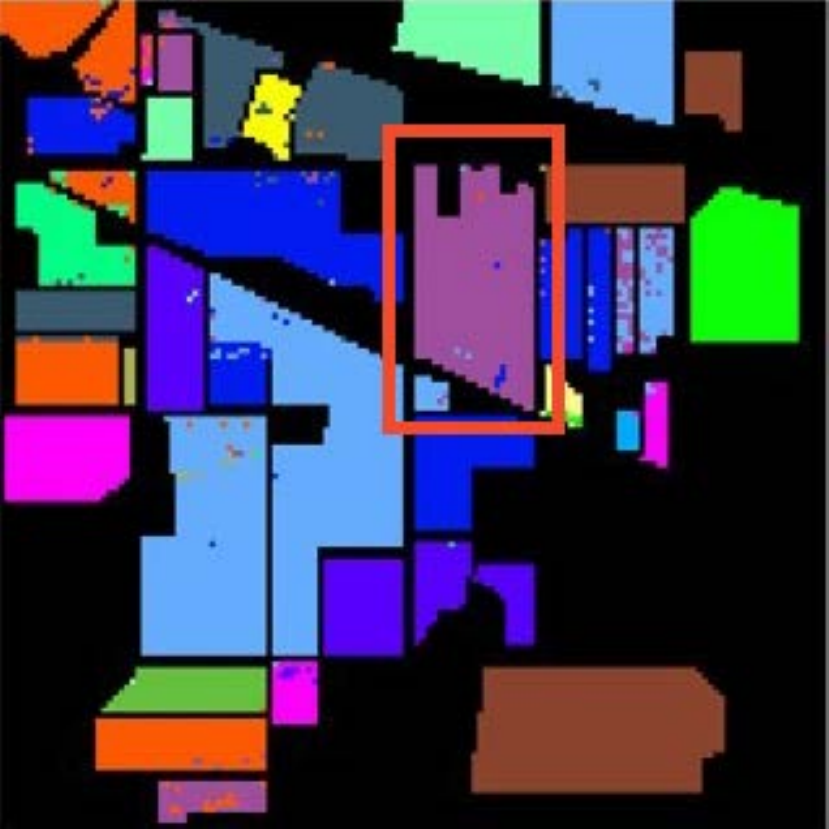}}
	\subfigure[]{\includegraphics[width=2.8cm,height = 2.8cm]{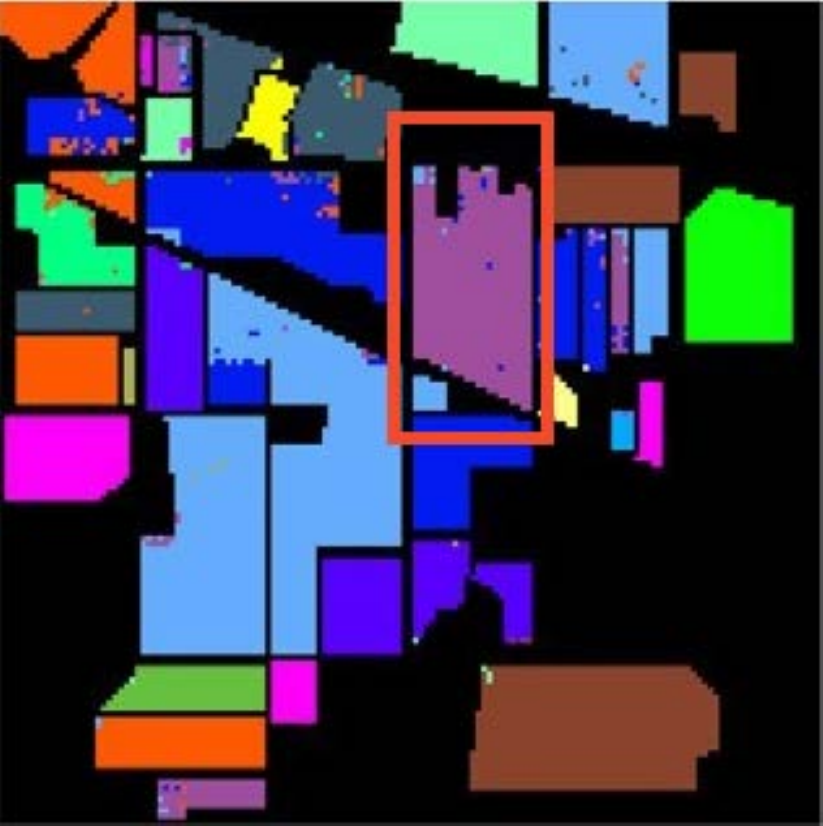}}
	\subfigure[]{\includegraphics[width=2.8cm,height = 2.8cm]{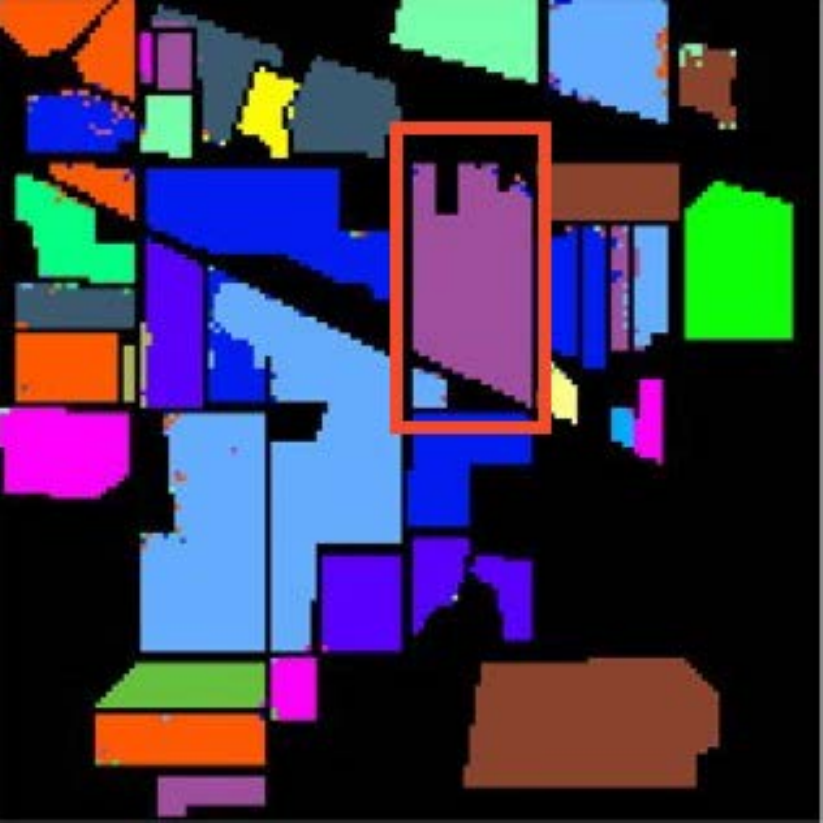}}
	\caption{ Classification results for Indian Pines dataset (5 percent of samples). (a) 3D-CNN (b) CCF (c) LDCR (d) SSRN (e) MSDN-SA (f) FCSPN}
	\label{fig:camoflage_detection}
	\vspace{-0.25cm}
\end{figure}

\begin{figure}[htp]
	\centering
	\setlength{\abovecaptionskip}{0pt}
	\subfigure[]{\includegraphics[width=2.8cm,height = 2.8cm]{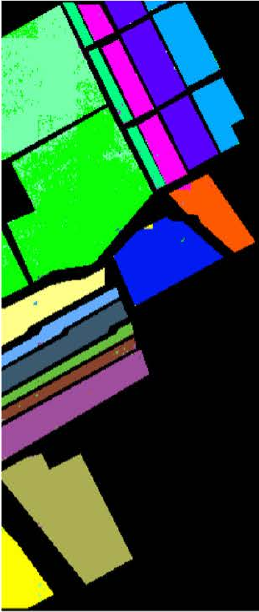}}
	\subfigure[]{\includegraphics[width=2.8cm,height = 2.8cm]{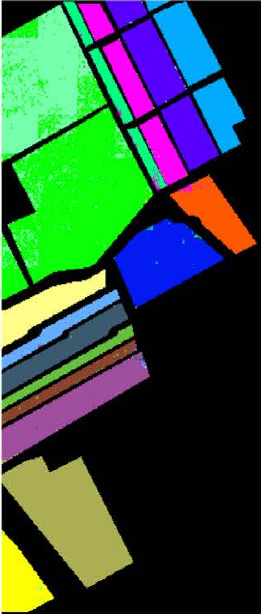}}
	\subfigure[]{\includegraphics[width=2.8cm,height = 2.8cm]{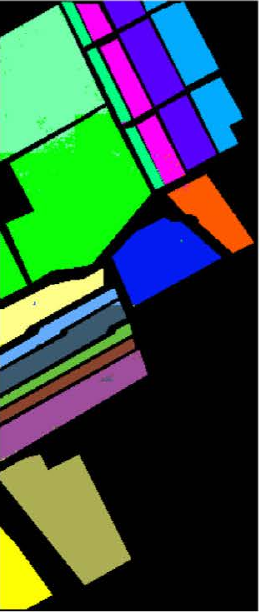}}
	\subfigure[]{\includegraphics[width=2.8cm,height = 2.8cm]{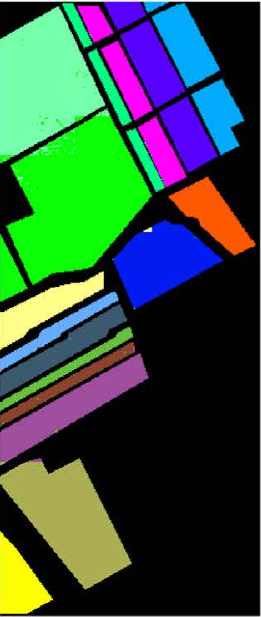}}
	\subfigure[]{\includegraphics[width=2.8cm,height = 2.8cm]{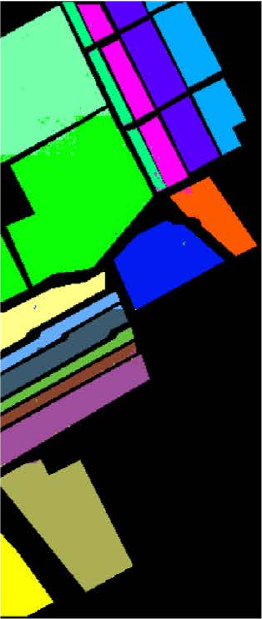}}
	\subfigure[]{\includegraphics[width=2.8cm,height = 2.8cm]{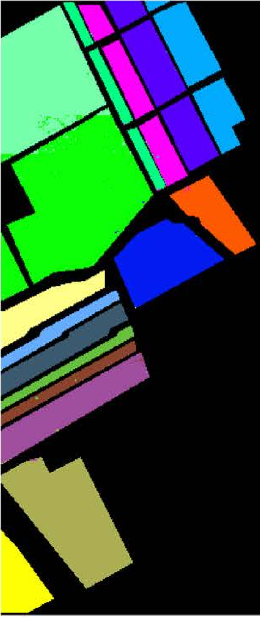}}
	\caption{ Classification results for Salinas dataset (5 percent of samples). (a) CCF (b) 3D-CNN (c) LDCR (d) SSRN (e) MSDN-SA (f) FCSPN}
	\label{fig:camoflage_detection}
	\vspace{-0.25cm}
\end{figure}

\begin{figure*}[htp]
	\centering
	\setlength{\abovecaptionskip}{0pt}
	\subfigure[]{\includegraphics[width=16.8cm,height = 2.5cm]{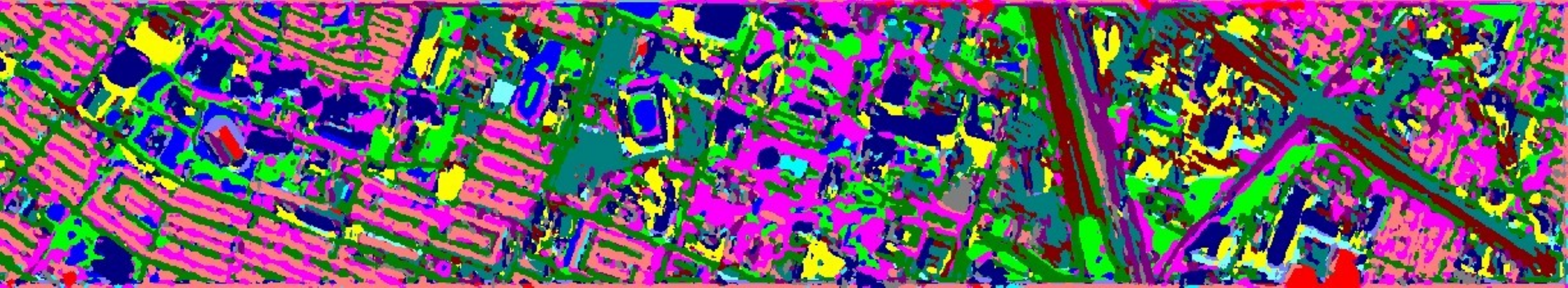}}
	\subfigure[]{\includegraphics[width=16.8cm,height = 2.5cm]{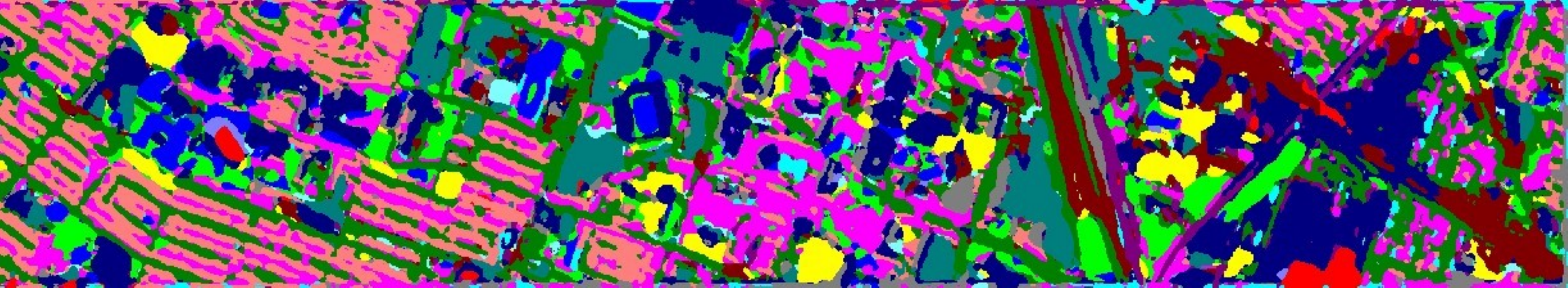}}
	\subfigure[]{\includegraphics[width=16.8cm,height = 2.5cm]{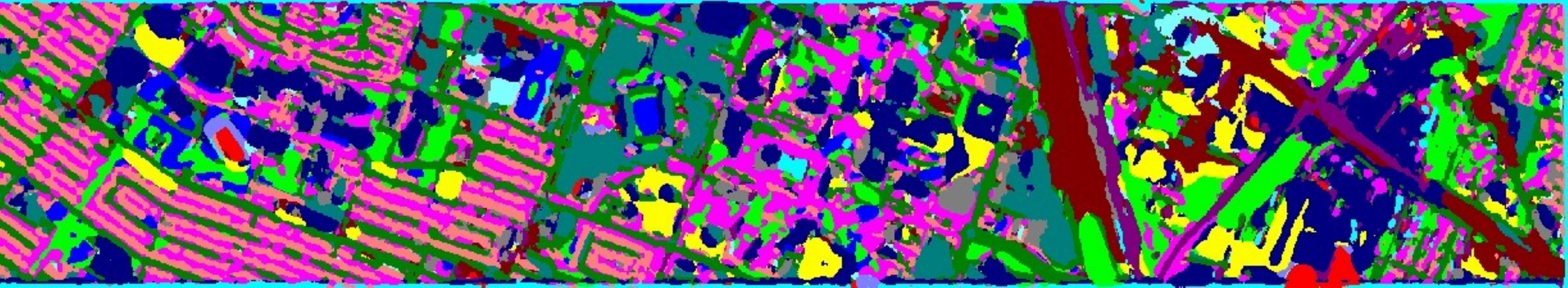}}
	\subfigure[]{\includegraphics[width=16.8cm,height = 2.5cm]{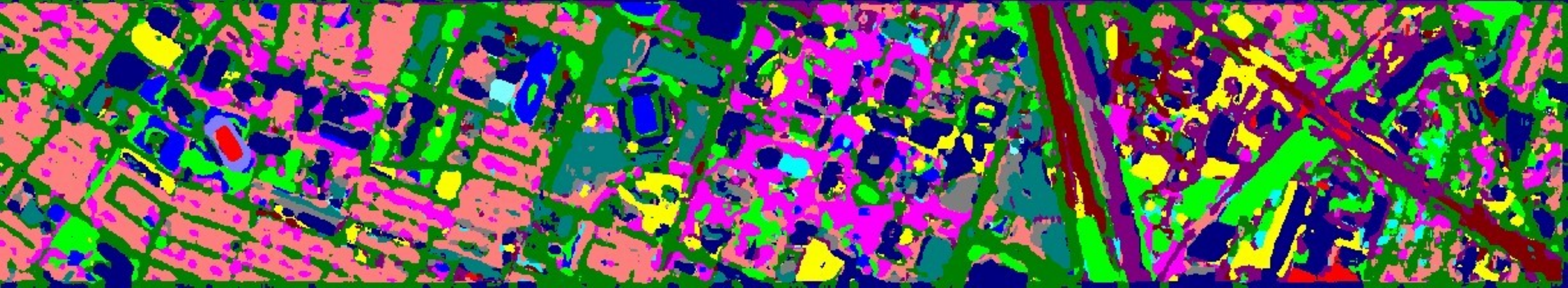}}
	\subfigure[]{\includegraphics[width=16.8cm,height = 2.5cm]{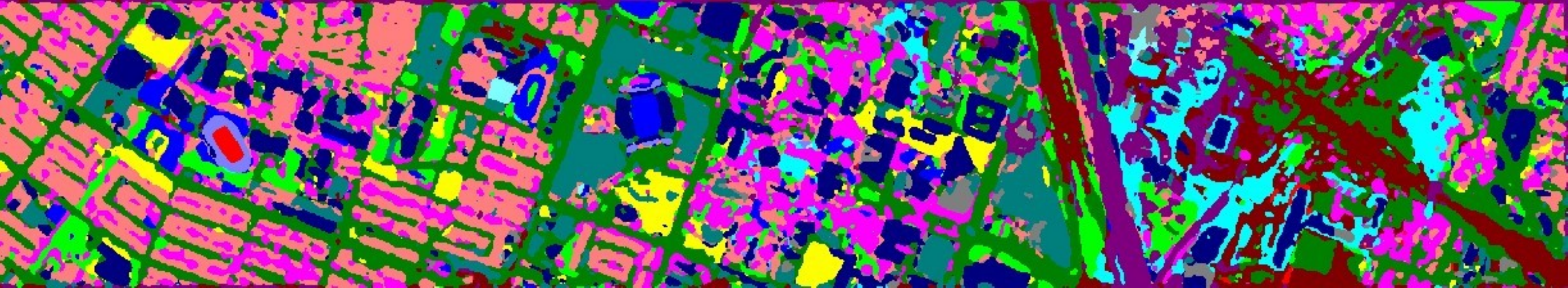}}
	\subfigure[]{\includegraphics[width=16.8cm,height = 2.5cm]{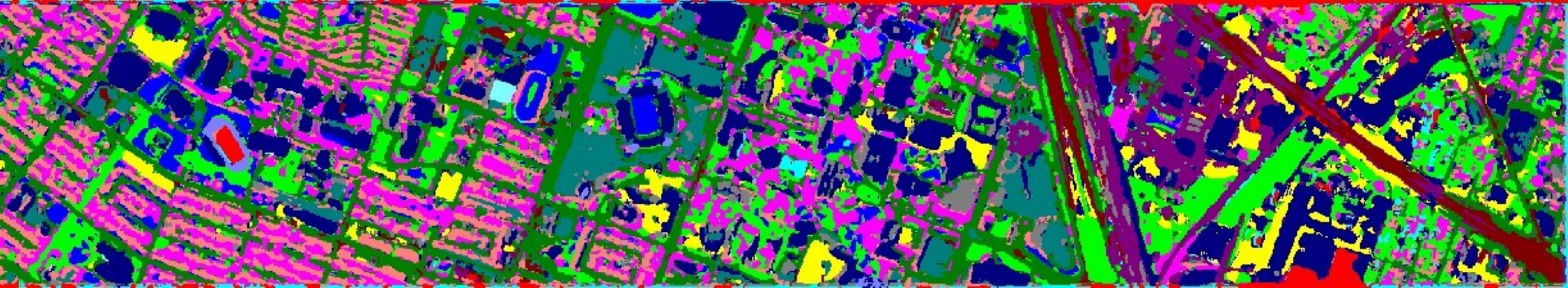}}
	\caption{ Classification results for University of Houston (5 percent of samples). (a) 3D-CNN (b) CCF (c) LDCR (d) MSDN-SA (e) SSRN (f) FCSPN}
	\label{fig:camoflage_detection}
\end{figure*}

The third dataset is the University of Houston, which was gathered in 2012 by the National Airborne Laser Mapping Center (NCALM) funded by the National Science Foundation deploying the ITRES-CASI (Compact Airborne Spectral Imager) 1500 hyperspectral Imager on the University of Houston campus and the neighboring urban areas. Fifteen land-cover classes with 349 pixels $\times$ 1905 pixels, and 144 bands with wavelength ranging from 0.36 to 1.05 $\mu m$ were included. Fig. 7 illustrates the false color composite of the University of Houston image and the corresponding ground reference map, respectively.

\subsection{Implementation Details}
In the experimental setting, the sampling strategy highlights its functionality role in the classifier learning and evaluation~\cite{Zhou2015On}. In our experiments, two sampling strategies are deployed to create the training sets. The first is to randomly select 200 samples for each class, and alternatively randomly collect 5 percent of samples from the whole reference image.
We adopt the focal loss~\cite{lin2017focal} as the loss function. 
%In order to overcome sample imbalance in HSI, L2 is added regularization term to the focal loss to restrict the sum of the squares of the parameters to be small. 
In order to overcome the sample imbalance in HSI, the L2 regularization term is added to the focus loss to limit the parameter square sum to be small.
Stochastic gradient descent (SGD) is adopted as the optimizer~\cite{krizhevsky2012imagenet}, where batch size, weight decay, the number of training epochs, momentum, and learning rate are set to 20, 1e-5, 60, 0.9, and 0.01, respectively. 
%where momentum, weight decay, batch size, the number of training epochs, and learning rate are set to 0.9, 1e-5, 20, 60, and 0.01, respectively. 
Quantitative metrics are overall accuracy (OA), average accuracy (AA), and Cohen’s Kappa statistic ($\kappa$), where OA measures the probability of consistency between the prediction results for each pixel; AA emphasizes on the class level, which represents the average percentage of pixels correctly classified in each class, and $\kappa$ estimates the accuracy of classification pixels corrected by the number of agreements expected purely by chance. 

\subsection{Comparison with the state-of-the-art}
\begin{table*}[htp]
	
	\setlength{\abovecaptionskip}{0pt}
	
	\caption{Network architecture details of proposed FCSPN for Salinas Dataset (200 samples per class)}
	
	\centering
	
	\scalebox{1.16}{
		
		\begin{tabular}{c|c|c|c|c|c|c}
			
			\hline
			
			\multirow{2}{*}{Class} & \multicolumn{6}{c}{Methods}                                                                                                                                              \\ \cline{2-7} 
			
			& \multicolumn{1}{c|}{CCF~\cite{xia2017hyperspectral}} & \multicolumn{1}{c|}{3D-CNN~\cite{Li2017Spectral}} & \multicolumn{1}{c|}{SSRN~\cite{zhong2017spectral}} & \multicolumn{1}{c|}{LDCR~\cite{zhang2019learning}} & \multicolumn{1}{c|}{MSDN-SA~\cite{fang2019hyperspectral}}  & \multicolumn{1}{c}{FCSPN} \\ \hline
			
			Brocoli green weeds 1                      &           96.75               &         99.81                    & 99.75        &99.79                  &       99.97                                      &    100                        \\ 
			
			Brocoli green weeds 2                      &             97.96             &           99.91                  & 99.86         &99.31                 &          99.84                          &  100                          \\ 
			
			Fallow                      &           98.02               &            98.36                 & 99.32           &99.67                &         99.97                            &  99.90                          \\ 
			
			Fallow rough plow                      &          98.62                &        99.37                     &99.5     &99.70                       &         99.85                               &   99.64                       \\ 
			
			Fallow smooth                      &         97.05                &         98.13                    & 99.65          &98.56                &         99.87                           &   99.81                         \\ 
			
			Stubble                      &        92.35                 &          85.09                   &  100              &99.79             &            100                                  &    100                        \\ 
			
			Celery                      &         96.72                 &         99.32                    & 100           &99.63                &          99.82                             &   100                         \\ 
			
			Grapes untrained                      &           85.64               &          91.89                   & 85.31         &76.83                 &          90.37                               &   99.40                         \\ 
			
			Soil vineyard develop                      &         90.25                 &         93.85                   &  94.58        &99.52                  &         99.90                                &     99.98                       \\ 
			
			Corn senesced green weeds                     &          95.75                &          97.99                   &99.38       &95.80                     &       98.90                                  &     100                       \\ 
			
			Lettuce romaine 4wk                     &          96.78                &         98.04                   &  96.66            &99.26            &     99.98                                           &   100                         \\ 
			
			Lettuce romaine 5wk                      &         93.28                 &         95.07                    &  96.85      &99.98                &         100                                    &    100                        \\ 
			
			Lettuce romaine 6wk                     &          80.25                &          77.08                   & 93.68       &99.75                 &       99.94                                      &   100                         \\ 
			
			Lettuce romaine 7wk                     &          95.29               &           97.52                 & 98.46         &98.92                 &      99.03                                    &    99.63                       \\ 
			
			Vineyard untrained                     &           76.32               &      78.50                       &  99.58           &79.72              &       93.06                                    &    98.46                        \\ 
			
			Vineyard vertical trellis                     &           93.26               &       95.84                      & 99.11       &98.77                   &           100                                &     99.67                       \\ \hline
			
			OA(\%)                 &       89.28                   &       92.30                      & 93.16                 &91.58        &        96.81                                       &    {\bf99.63}                       \\ 
			
			AA(\%)                 &        90.56                  &     92.18                        &93.59            &96.56            &       98.78                                       &  {\bf 99.78}                         \\ 
			
			$\kappa \times 100$                  &       90.35                  &        91.85                     &  93.16            &90.59               &    98.55                                        &    {\bf99.58}                        \\ \hline
			
		\end{tabular}
		
	}
	
	\label{salinas_200}	
\end{table*}

\begin{table*}[htp]
	
	\setlength{\abovecaptionskip}{0pt}
	
	\caption{Network architecture details of proposed FCSPN for Salinas Dataset (5 percent of samples)}
	
	\centering
	
	\scalebox{1.16}{
		
		\begin{tabular}{c|c|c|c|c|c|c}
			
			\hline
			
			\multirow{2}{*}{Class} & \multicolumn{6}{c}{Methods}                                                                                                                                              \\ \cline{2-7} 
			
			& \multicolumn{1}{c|}{CCF~\cite{xia2017hyperspectral}} & \multicolumn{1}{c|}{3D-CNN~\cite{Li2017Spectral}} & \multicolumn{1}{c|}{SSRN~\cite{zhong2017spectral}} & \multicolumn{1}{c|}{LDCR~\cite{zhang2019learning}}& \multicolumn{1}{c|}{MSDN-SA~\cite{fang2019hyperspectral}} & \multicolumn{1}{c}{FCSPN} \\ \hline
			
			Brocoli green weeds 1                      &           95.65               &         98.19                    & 100      &99.45                    &       99.72                                     &    100                        \\ 
			
			Brocoli green weeds 2                     &             96.77             &           99.81                  & 100        &99.92                  &          99.87                               &  100                          \\ 
			
			Fallow                     &           95.74               &            96.38                 & 98.31        &99.49                  &         99.84                            &  99.29                          \\ 
			
			Fallow rough plow                       &          96.32                &        98.73                     &99.32        &99.21                  &         99.00                               &   99.78                         \\ 
			
			Fallow smooth                     &         96.34                 &         98.01                    & 99.46        &99.44                  &         99.79                                  &   99.59                         \\ 
			
			Stubble                      &        96.46                  &          98.79                   &  100           &100              &            99.97                                  &    100                        \\ 
			
			Celery                      &         95.98                 &         98.01                    & 100           &99.08               &          99.78                                 &   100                         \\ 
			
			Grapes untrained                      &           80.23               &          84.90                   & 93.62           &92.56               &          97.38                                    &   99.15                         \\ 
			
			Soil vineyard develop                      &         97.23                 &         98.23                    &  98.90         &99.94               &         100                            &     99.97                       \\ 
			
			Corn senesced green weeds                     &          88.45                &          91.84                   &99.32       &95.42                   &       99.13                                   &     99.88                       \\ 
			
			Lettuce romaine 4wk                     &          92.14                &         93.58                    &  98.73              &97.38           &     99.62                                  &   97.47                         \\ 
			
			Lettuce romaine 5wk                     &         95.92                 &         97.29                    &  100            &100             &        100                                       &    100                        \\ 
			
			Lettuce romaine 6wk                     &          94.10                &          98.01                   & 100              &99.56            &       99.44                                &   100                         \\ 
			
			Lettuce romaine 7wk                     &          94.92                &           95.02                 & 98.99           &97.29               &      97.04                                           &    98.22                        \\ 
			
			Vineyard untrained                     &           84.69               &      75.80                       &  99.32           &86.06              &       95.93                                  &    94.41                        \\ 
			
			Vineyard vertical trellis                     &           95.78               &       96.57                      & 100          &97.34               &           99.70                               &     98.06                       \\ \hline
			
			OA(\%)                 &       90.81                   &       92.30                      & 97.96             &95.94             &        98.70                                       &    {\bf98.86}                       \\ 
			
			AA(\%)                 &        92.56                  &     93.18                        &98.41               &97.63            &       98.14                                    &  {\bf 99.11}                         \\ 
			
			$\kappa \times 100$                  &       91.58                   &        92.85                     &  98.01                &95.48         &    98.55                                          &    {\bf98.73}                        \\ \hline
			
		\end{tabular}
		
	}
	
	\label{salinas_005}	
\end{table*}

\begin{table*}[htp]
	
	\setlength{\abovecaptionskip}{0pt}
	
	\caption{Network architecture details of proposed FCSPN for University of Houston Dataset (200 samples per class)}
	
	\centering
	
	\scalebox{1.28}{
		
		\begin{tabular}{c|c|c|c|c|c|c}
			
			\hline
			
			\multirow{2}{*}{Class} & \multicolumn{6}{c}{Methods}                                                                                                                                              \\ \cline{2-7} 
			
			& \multicolumn{1}{c|}{CCF~\cite{xia2017hyperspectral}} & \multicolumn{1}{c|}{3D-CNN~\cite{Li2017Spectral}} & \multicolumn{1}{c|}{SSRN~\cite{zhong2017spectral}} & \multicolumn{1}{c|}{LDCR~\cite{zhang2019learning}} & \multicolumn{1}{c|}{MSDN-SA~\cite{fang2019hyperspectral}} & \multicolumn{1}{c}{FCSPN} \\ \hline
			
			Healthy grass                      &     80.25                     &       82.16                    &     96.93        &86.87            &          92.60                                  &       98.96                     \\ 
			
			Stressed grass                    &   83.26                       &       85.19                     &      96.94           &86.24          &   94.21                                        &         99.68                   \\ 
			
			Synthetic grass                     &  85.70                        &      90.69                       &    98.52          &93.96              &   98.23                                  &         99.57                   \\ 
			
			Trees                      &  90.91                        &    93.35                         &     96.20       &65.71               &   85.67                                          &         99.84                \\ 
			
			Soil                      &    93.36                      &      99.61                       &   98.53          &92.80             &  97.52                                      &          99.76               \\ 
			
			Water                     &  94.36                        &      97.89                       &  96.90         &81.60                &    96.64                              &        100                 \\ 
			
			Residential                      & 90.96                         &      91.25                       &  98.29       &61.80                  & 86.40                                    &       98.82                \\ 
			
			Commercial                     &   70.85                       &   75.92                         &      98.58       &81.70          &   91.57                                &           98.71               \\ 
			
			Road                     &78.44                          &    83.20                        &  98.02             &69.58           &   85.51                               &           98.32                 \\ 
			
			Highway                    &   60.85                       &    48.80                        &   98.65           &91.63             &  93.85                                   &          100                 \\ 
			
			Railway                   &  65.32                        &   69.21                         &    97.03        &79.52                &   94.72                               &         100                 \\ 
			
			Parking Lot 1                    & 82.97                         &    85.55                       &    95.99          &83.83              &   92.04                             &          99.11                 \\ 
			
			Parking Lot 2                     &  85.88                        &   90.85                         &    97.25          &81.04             &     96.95                             &          99.79                \\ 
			
			Tennis Court                     &  87.43                        &      90.66                       &      98.56        &95.61            &    99.82                               &         100                 \\ 
			
			Running Track                    &   78.65                       &      76.83                       &  97.30          &90.43              &  97.04                                &         99.70                   \\ \hline
			
			OA(\%)                 &   79.44                       &      82.25                       &   85.12         &81.23             &  92.22                                   &         {\bf99.39}                   \\ 
			
			AA(\%)                 &    83.64                      &  84.08                         &   85.47          &82.82             &   93.52                                 &   {\bf99.48}                         \\ 
			
			$\kappa \times 100$                  &    79.65                      &      80.13                     &      83.90           &79.64          &   91.56                             &    {\bf99.35}                        \\ \hline
			
		\end{tabular}
		
	}
	
	\label{IG13_200}

\end{table*}

\begin{table*}[htp]
	
	\setlength{\abovecaptionskip}{0pt}
	
	\caption{Network architecture details of proposed FCSPN for University of Houston Dataset (5 percent of samples)}
	
	\centering
	
	\scalebox{1.28}{
		
		\begin{tabular}{c|c|c|c|c|c|c}
			
			\hline
			
			\multirow{2}{*}{Class} & \multicolumn{6}{c}{Methods}                                                                                                                                              \\ \cline{2-7} 
			
			& \multicolumn{1}{c|}{CCF~\cite{xia2017hyperspectral}} & \multicolumn{1}{c|}{3D-CNN~\cite{Li2017Spectral}} & \multicolumn{1}{c|}{SSRN~\cite{zhong2017spectral}}  & \multicolumn{1}{c|}{LDCR~\cite{zhang2019learning}}& \multicolumn{1}{c|}{MSDN-SA~\cite{fang2019hyperspectral}} & \multicolumn{1}{c}{FCSPN} \\ \hline
			
			Healthy grass                      &     85.76                     &         90.56                    &     85.73        &79.46               &         91.63                                     &       91.77                     \\ 
			
			Stressed grass                     &   79.30                       &       68.52                      &      94.47        &68.66             &   90.26                                          &         91.71                   \\ 
			
			Synthetic grass                     &  82.07                        &      90.86                       &    98.47         &89.75               &    96.81                               &         99.28                   \\ 
			
			Trees                      &  70.01                        &    64.41                         &     92.22              &67.28         &   85.51                                          &         90.76                   \\ 
			
			Soil                      &    90.43                      &      92.87                       &   99.57              &88.57           &  97.81                                   &          99.03                  \\ 
			
			Water                     &  70.20                        &      65.91                       &  100              &47.69            &    79.37                              &        87.38                    \\ 
			
			Residential                     & 50.96                         &      65.84                       &  79.64             &45.66             & 81.73                                 &         86.83                   \\ 
			
			Commercial                     &   65.59                       &    56.19                         &      96.09            &72.99          &    85.80                                   &           91.24                 \\ 
			
			Road                     &65.95                          &    78.45                         &  86.20               &54.07           &   81.68                                     &           89.30                 \\ 
			
			Highway                     &   78.41                       &     58.48                        &   92.57           &79.71              &  93.41                                     &          96.82                  \\ 
			
			Railway                    &  63.94                        &   48.96                          &    99.03           &65.83             &   94.31                                     &         98.87                   \\ 
			
			Parking Lot 1                     & 72.17                         &     69.70                        &    89.14           &79.48             &    93.58                                     &          96.92                  \\ 
			
			Parking Lot 2                     &  65.10                        &   54.97                          &    96.27           &42.22            &     85.72                              &          97.44                  \\ 
			
			Tennis Court                    &  89.34                        &      86.91                       &      96.19        &72.20             &    94.93                                    &         98.60                   \\ 
			
			Running Track                     &   68.43                       &      86.39                       &  96.96             & 76.36            &  89.52                                &         100                   \\ \hline
			
			OA(\%)                 &   68.90                       &      67.61                       &    91.98        &69.90                & 89.69                                        &         {\bf94.02}                   \\ 
			
			AA(\%)                 &    69.34                      &   68.77                          &   92.50          &68.54               &  89.47                                       &   {\bf94.40}                         \\ 
			
			$\kappa \times 100$                  &    63.21                      &       65.38                      &      91.33        &67.47             &    88.85                                       &    {\bf93.53}                        \\ \hline
			
		\end{tabular}
		
	}
	
	\label{IG13_005}

\end{table*}
In our experiments, five state-of-the-art HSI classification methods will be evaluated to compare with the proposed FCSPN method. All the comparison methods are summarized as follows:
\begin{itemize}
  \item [1)] 
  CCF~\cite{xia2017hyperspectral}: Canonical Correlation Forests based on spectral feature with 100 trees.      
  \item [2)]
  3D-CNN~\cite{Li2017Spectral}:  Two 3D convolutional layers and a fully connected layer are contained in the 3D-CNN network framework. The setting of network structure is described in~\cite{Li2017Spectral}.
   \item [3)]
  SSRN~\cite{zhong2017spectral}: In~\cite{zhong2017spectral}, the architecture of the SSRN was illustrated detailedly. Two convolutional layers and two spectral residual blocks are included in the spectral feature learning part, while the spatial feature learning part consists of one 3D convolutional layers and two spatial residual blocks. Eventually, an average pooling layer and a fully connected layer are adopted to output the results.
   \item [4)]
   LDCR~\cite{zhang2019learning}: The LDCR develops a multi-task deep learning framework, which incorporates classification and HSIs compression into one framework to learn the discriminative compact representation of HSI classification. 
  \item [5)]
  MSDN-SA~\cite{fang2019hyperspectral}: In MSDN-SA, the spectral and spatial features were captured at different scales simultaneously by 3D dilated convolutions, and all 3D feature maps are connected densely. Besides, a spectral-wise attention mechanism is introduced to improve the distinguishability of spectral features. 
\end{itemize}

Below we compare the above algorithms with our proposed FCSPN. For a fair comparison, the training sample of the three HSI datasets was set to the same amount .
\subsubsection{Results of Indian Pines Dataset}
In the first experiment, we follow the first sampling strategy to randomly select 200 samples (Except for the classes of Alfalfa, Grass-pasture-mowed, Oats and Stone-steel-towers class, which total sample size are less than 200. Training samples of these four classes are 10, 6,10, 9, respectively.) from each class of reference map for training, while the rest of the samples are used for testing. Table~\ref{In_200} gives numerical evaluation results on the classification performance. 
Taken together from the table, these results suggest that the classification performance is improved due to utilizing the FCSPN, which proves the positive effect of spatial consistency on HSI classification. 
In the second experiment of the Indian Pines dataset, we employ the second sampling strategy that randomly select 5$\%$ of the samples from the ground-truth reference for training purpose the rest of samples are used for validation. 
%The average classification accuracies and the corresponding standard deviations of the Indian Pines dataset are reported in Table~\ref{In_005} and the classification maps of different methods are presented in Fig.8. 
Table~\ref{In_005} reports the average classification accuracies and the corresponding standard deviations of Indian Pines dataset, while Fig.8 presents the classification maps of different approaches.
When the second sampling strategy is adopted for this dataset, MSDN-SA and FCSPN are superior to other approaches, and the advantage of FCSPN is approximately 3$\%$ to 20$\%$. 
%Note that the numbers of class samples in this dataset are quite unbalanced. In particular, those of the classes Alfalfa, Grass-pasture-mowed, Oats and Stone-steel-towers are very few. Due to focal loss, our proposed method has achieved the highest classification accuracy.

\subsubsection{Results of Salinas Dataset}
Table~\ref{salinas_200} and Table~\ref{salinas_005} respectively show the experimental results of sample selection by the first sampling strategy (200 samples for each class) and the second one (5 percent of samples from the whole reference image).
As shown in Table~\ref{salinas_200}, when we adopt the first sampling strategy, our FCSPN has a significant improvement in the quantitative metrics OA, AA, and $\kappa$. 
Meanwhile, Table~\ref{salinas_005} demonstrates that the quantitative metrics are also slightly improved when the sample selection method of strategy two is selected. Besides, Fig.9 indicates the classification maps gained by different approaches.
Regardless of the employed selection methods, FCSPN can be found to stand out from other competitors, according to OA, AA, and $\kappa$.

\subsubsection{Results of University of Houston Dataset}
The the experimental results of the University of Houston dataset are demonstrated in Table~\ref{IG13_200} and Table~\ref{IG13_005}, separately. 
It can be seen in Table~\ref{IG13_200} that under first sampling strategy (200 samples for each class) is adopted, the quantitative metrics are significantly improved. 
In addition to visual comparison, Table IV displays the quantitative results of six methods on the image, in which OA, AA, and $\kappa$ are adopted to evaluate the classification performance.
Specifically, OA, AA and $\kappa$ are increased by 7.17$\%$, 5.96$\%$ and 7.79$\%$, respectively.
Simultaneously, Fig.10 shows the classification maps gained by different methods. 
As can be seen from this figure, it states that the classification maps of FCSPN appear to be more smoothing and more detailed in edges. 
In general, the advantage of FCSPN for this dataset is that the performance is more remarkable compared to the other five algorithms. Our proposed FCSPN yields the highest OA, AA, and $\kappa$ and is significantly superior to other competitors.

\subsection{Effect of Training Samples}
As shown in Fig.11, Fig.12 and Fig.13, we randomly chose \{50,100,150,200\} samples from each class of Indian Pines, Salinas and University of Houston for training. 
Notably, as the number of training samples increases, the accuracy of OA, AA and $\kappa$ improved obviously. 
When the training sample reached 200, the improvement in these coefficients was negligible. Therefore, we choose 200 as the final number of training samples.

\begin{figure}[htp]
 \centering
 \includegraphics[scale=0.35]{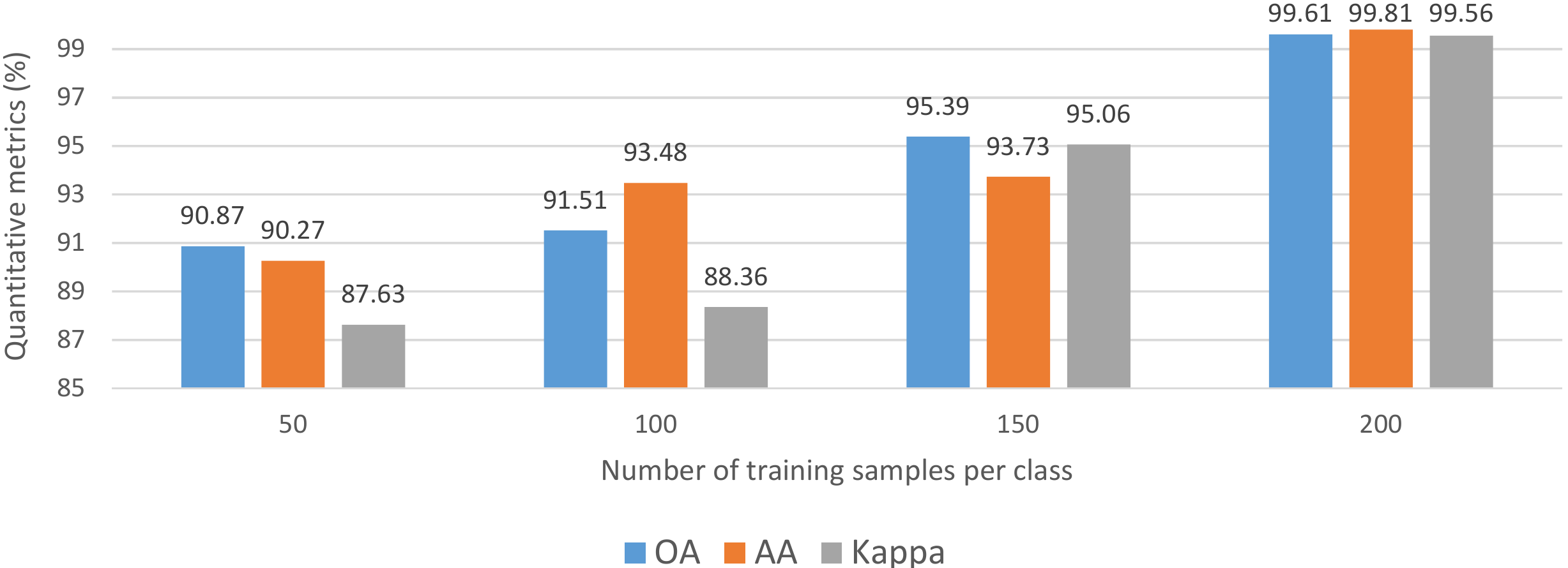}
 \caption{Classification accuracy ($\%$) of the proposed algorithm for Indian Pines with 50,100,150 and 200 training samples per class.}
 \label{patch_size}
\vspace{-0.25cm}
\end{figure}

\begin{figure}[htp]
 \centering
 \includegraphics[scale=0.35]{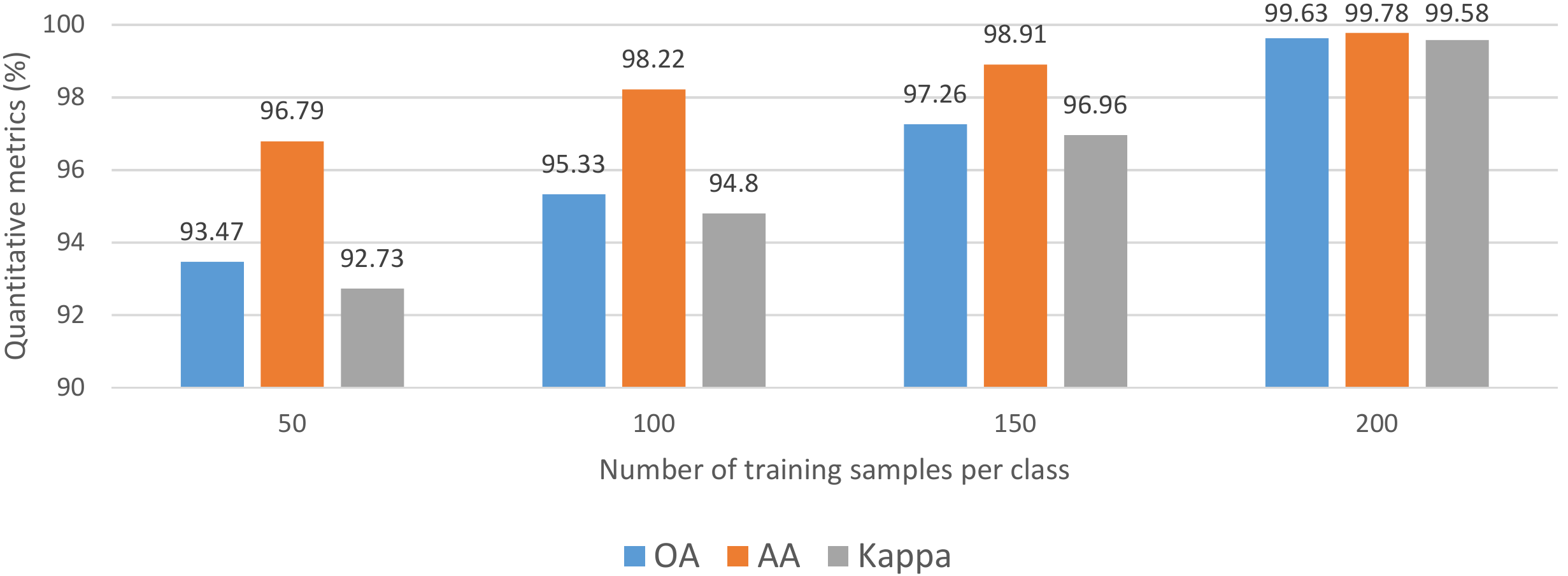}
 \caption{Classification accuracy ($\%$) of the proposed algorithm for Salinas with 50,100,150 and 200 training samples per class.}
 \label{patch_size}
\vspace{-0.25cm}
\end{figure}

\begin{figure}[htp]
 \centering
 \includegraphics[scale=0.35]{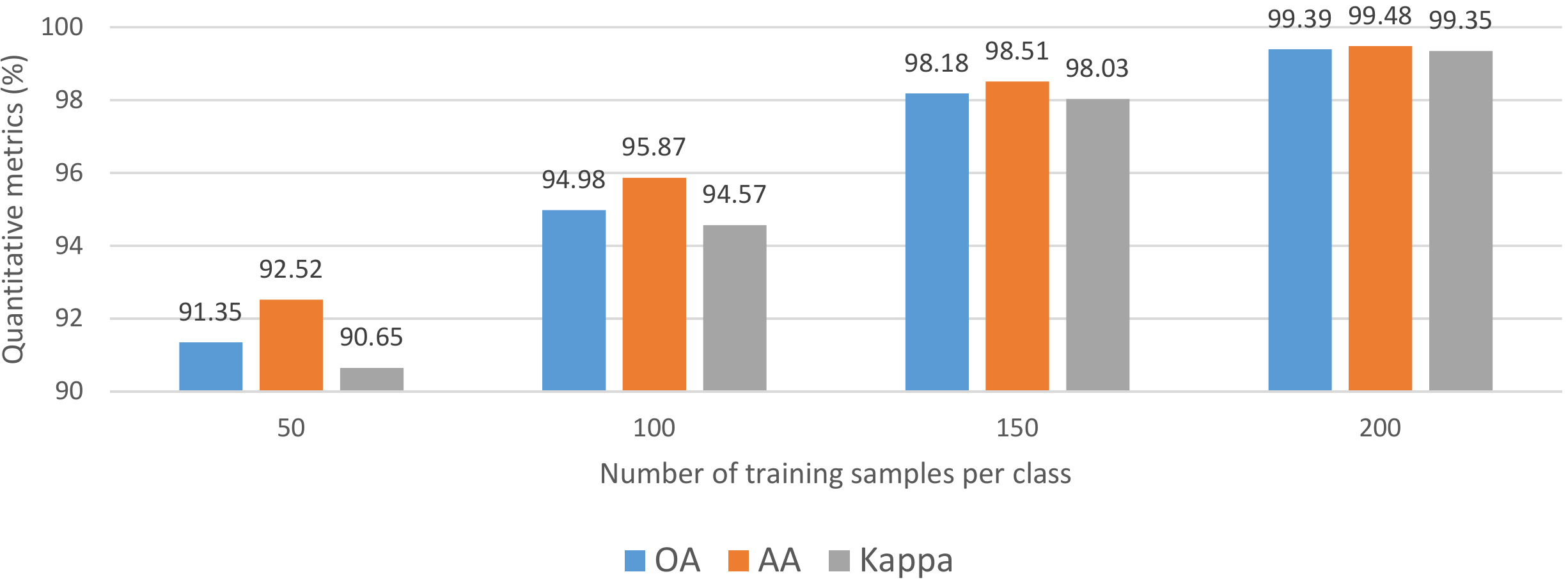}
 \caption{Classification accuracy ($\%$) of the proposed algorithm for University of Houston with 50,100,150 and 200 training samples per class.}
 \label{patch_size}
\vspace{-0.25cm}
\end{figure}

\subsection{Ablation Study for CSPN}
\begin{table}[]
\setlength{\abovecaptionskip}{0pt}
 \caption{ Ablation Study of Indian Pines Dataset}
 \centering
 \begin{tabular}{c|c|c|c|l}
  \hline
  \multicolumn{2}{c|}{Quantitative metrics}          & OA(\%)  & AA(\%)  & $\kappa \times 100$  \\ \hline
  \multirow{2}{*}{200 samples per class} & 3D-FCN   &   99.24  &  99.67   &  99.13 \\ \cline{2-5} 
  & FCSPN & {\bf99.61}   & {\bf99.81}   & {\bf99.56}  \\ \hline
  \multirow{2}{*}{5 percent of samples}        & 3D-FCN   & 97.36   & 97.04   &  96.99 \\ \cline{2-5} 
  & FCSPN &   {\bf98.22}  &  {\bf97.84}   &  {\bf98.22}  \\ \hline
 \end{tabular}
 \label{AS_In}
\end{table}

\begin{table}[]
\setlength{\abovecaptionskip}{0pt}
 \caption{ Ablation Study of Salinas Dataset}
 \centering
 \begin{tabular}{c|c|c|c|l}
  \hline
  \multicolumn{2}{c|}{Quantitative metrics}          & OA(\%)  & AA(\%)  & $\kappa \times 100$  \\ \hline
  \multirow{2}{*}{200 samples per class} & 3D-FCN   &   97.45  &  98.83   &   97.16 \\ \cline{2-5} 
  & FCSPN & {\bf99.63}   & {\bf99.78}   & {\bf99.58}  \\ \hline
  \multirow{2}{*}{5 percent of samples}        & 3D-FCN   & 98.42   & 98.75   &  98.24 \\ \cline{2-5} 
  & FCSPN &   {\bf98.86}  &  {\bf99.11}   &  {\bf98.73}  \\ \hline
 \end{tabular}
 \label{AS_Sa}
\end{table}

\begin{table}[]
\setlength{\abovecaptionskip}{0pt}
 \caption{ Ablation Study of University of Houston Dataset}
 \centering
 \begin{tabular}{c|c|c|c|l}
  \hline
  \multicolumn{2}{c|}{Quantitative metrics}          & OA(\%)  & AA(\%)  & $\kappa \times 100$  \\ \hline
  \multirow{2}{*}{200 samples per class} & 3D-FCN   &  97.52  & 97.77    &   97.32 \\ \cline{2-5} 
  & FCSPN & {\bf99.39}   & {\bf99.48}   & {\bf99.35}  \\ \hline
  \multirow{2}{*}{5 percent of samples}        & 3D-FCN   & 92.01   & 92.63   &  91.36 \\ \cline{2-5} 
  & FCSPN &   {\bf94.02}  &  {\bf94.40}   &  {\bf93.53}  \\ \hline
 \end{tabular}
 \label{AS_IG13}
\end{table}
In order to prove the significance of CSPN, we also conduct ablation experiments on three datasets, and the experimental results are listed in Table~\ref{AS_In}, Table~\ref{AS_Sa} and Table~\ref{AS_IG13}.
These three tables can represent the superiority of FCSPN with two sample selection strategies, which indicates the proposed FCSPN method performs higher classification accuracy than the 3D-FCN based method.

As shown in Table~\ref{AS_In}, since the accuracy of the 3D-FCN method on the Indian Pines dataset has reached 99$\%$, the FCSPN method has slightly improved the accuracy with the refinement of CSPN under 200 samples for each class. While we employ the second sampling strategy (5 percent of samples from the whole reference image), the classification accuracy has improved more dramatically.

In Salinas dataset, the first sampling strategy, the evaluation indexes of FCSPN method are improved higher than that of the second sampling strategy, comparatively, as indicated in Table~\ref{AS_Sa}.

Table~\ref{AS_IG13} demonstrates that regardless of sampling strategy is adopted in dataset University of Houston, the FCSPN method proposed by us has a significant improvement in classification accuracy compared with the 3D-FCN based method.
\section{Conclusion}
In this paper, we have introduced a novel end-to-end, pixel-to-pixel approach for HSI classification named FCSPN. 
Our proposed FCSPN architecture contains two efficient components. First, the 3D-FCN based preliminary classification method is adopted to extract HSI features in an end-to-end pipeline. 
A new DSR unit and channel-wise attention mechanism are employed, which effectively extracts both spectral and spatial feature information with fewer parameters and the most informative channels . 
The efficiency of feature extraction is improved significantly as the entire image can be input into the 3D-FCN based network for training once, which results in a reliable coarse classification result. 
Secondly, the deployment of CSPN-based refinement strategy contributes further to capturing spatial consistency information within HSIs and thus the classification performance can be refined.
%Second, the CSPN-based refinement strategy is proposed, which provides a robust method to describe the consistency spatial feature within HSIs, so that the classification accuracy can be further improved.
%Secondly, the refinement strategy based on CSPN is proposed, which provides a robust method to describe the consistent spatial features in HSI,
%the CSPN-based network is employed to refine the coarse classification results based on FCN. 
%To make up for the insensitivity of FCN-based network to local spatial information, the CSPN-based optimization network utilizes the spatial information of local 8 neighborhoods, which further improves the classification accuracy. 
%We further evaluate the usefulness of our method by comparing it with several state-of-the-art methods and achieve promising results.
Experimental results on three widely used HSI datasets demonstrate that our proposed FCSPN is capable of yielding superior performance compared with the current state-of-the-art approaches.

{\small
\bibliographystyle{IEEEtran}
\bibliography{egbib}
}
\end{document}